\documentclass{article}

    \usepackage[final]{neurips_2023}

\usepackage[utf8]{inputenc} 
\usepackage[T1]{fontenc}    
\usepackage{hyperref}       
\usepackage{url}            
\usepackage{booktabs}       
\usepackage{amsfonts}       
\usepackage{nicefrac}       
\usepackage{microtype}      
\usepackage{xcolor}         
\usepackage{multirow}
\usepackage{algorithm}
\usepackage{algpseudocode}
\usepackage{makecell}

\usepackage{graphicx}
\usepackage{wrapfig}

\setcitestyle{square,numbers}
\hypersetup{colorlinks,linkcolor={blue},citecolor={green},urlcolor={magenta}}

\title{
Visual Programming for \\Text-to-Image Generation and Evaluation
}

\author{%
      Jaemin Cho \qquad
      Abhay Zala \qquad
      Mohit Bansal \\
      UNC Chapel Hill\\
  \texttt{\{jmincho, aszala, mbansal\}@cs.unc.edu} \\
    {\tt \normalsize \href{https://vp-t2i.github.io}{https://vp-t2i.github.io}}
}

\newcommand{\objectskill}[1]{Object}
\newcommand{\countskill}[1]{Count}
\newcommand{\spatialskill}[1]{Spatial}
\newcommand{\scaleskill}[1]{Scale}
\newcommand{\textskill}[1]{Text Rendering}

\newcommand{\sd}[1]{Stable Diffusion}
\newcommand{\sdcompvis}[1]{\sd{} v1.4}
\newcommand{\sdstabai}[1]{\sd{} v2.1}

\newcommand{\karlo}[1]{Karlo}
\newcommand{\dallemega}[1]{DALL-E Mega}
\newcommand{\mindalle}[1]{minDALL-E}

\newcommand{\gligen}[1]{GLIGEN}

\newcommand{\textboxmethod}{\textsc{VPGen}}

\newcommand{\vicuna}{Vicuna}

\newcommand{\tifaprompts}{TIFA160}

\newcommand{\eval}{\textsc{VPEval}}

\newcommand{\etc}{\textit{etc}}

\newcommand{\ie}{\textit{i}.\textit{e}.,}
\newcommand{\eg}{\textit{e}.\textit{g}.,}
\newcommand{\vs}{\textit{v.s.}}

\usepackage[capitalize]{cleveref}
\crefname{section}{Sec.}{Secs.}
\Crefname{section}{Section}{Sections}
\Crefname{table}{Table}{Tables}
\crefname{table}{Tab.}{Tabs.}

\begin{document}

\maketitle

\begin{abstract}
As large language models have demonstrated impressive performance in many domains,
recent works have adopted language models (LMs) as controllers of visual modules for vision-and-language tasks.
While existing work focuses on equipping LMs with visual understanding,
we propose two novel interpretable/explainable visual programming frameworks 
for text-to-image (T2I) generation and evaluation.
First, we introduce \textboxmethod{}, an interpretable step-by-step T2I generation framework
that decomposes T2I generation into three steps: object/count generation, layout generation, and image generation.
We employ an LM to handle the first two steps (object/count generation and layout generation),
by finetuning it on text-layout pairs.
Our step-by-step T2I generation framework provides stronger spatial control than end-to-end models, the dominant approach for this task.
Furthermore, we leverage the world knowledge of pretrained LMs, overcoming the limitation of previous layout-guided T2I works that can only handle predefined object classes.
We demonstrate that our \textboxmethod{} has improved control
in counts/spatial relations/scales of objects than state-of-the-art T2I generation models.
Second, we introduce \eval{}, an interpretable and explainable evaluation framework for T2I generation based on visual programming.
Unlike previous T2I evaluations with a single scoring model that is accurate in some skills but unreliable in others,
\eval{} produces evaluation programs that invoke a set of visual modules that are experts in different skills, and also provides visual+textual explanations of the evaluation results.
Our analysis shows that \eval{} provides a more human-correlated evaluation for skill-specific and open-ended prompts than widely used single model-based evaluation.
We hope that our work encourages future progress on interpretable/explainable generation and evaluation for T2I models.
\end{abstract}

\begin{figure}
    \centering
  \includegraphics[width=.92\linewidth]{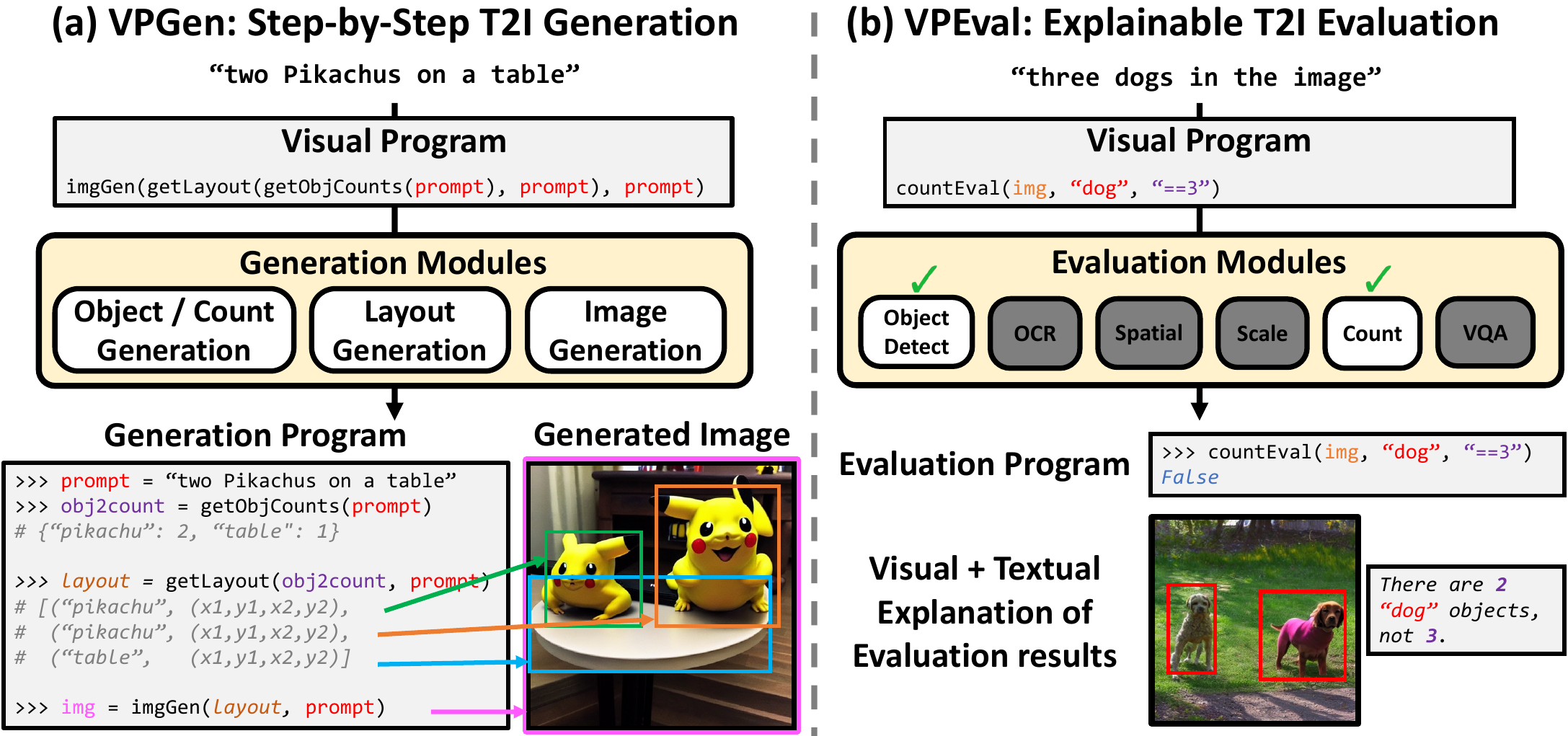}
  \caption{
  Illustration of the proposed visual programming frameworks for text-to-image (T2I) generation and evaluation.
  In (a) \textboxmethod{}, 
  we first generate a list of objects, then object positions, and finally an image, by executing three modules step-by-step.
  In (b) \eval{},
  we use evaluation programs with a mixture of evaluation modules that handle different skills and provide visual+textual explanation of evaluation results.
  }
\label{fig:teaser} 
\vspace{-12pt}
\end{figure}

\section{Introduction}
\label{sec:intro}

Large language models (LLMs) have shown remarkable performance on many natural language processing tasks, such as question answering, summarization, and story generation~\cite{Devlin2019,Lewis2020,Brown2020,Raffel2019,Chowdhery2022,Wei2022,OpenAI2023,Chung2022}.
Recent works have shown that LLMs can also tackle certain vision-and-language tasks 
such as visual question answering and visual grounding,
by generating visual programs that can control external visual modules and combine their outputs to get the final response~\cite{Yang2022PICa,Hu2022PromptCap,Gupta2023VisProg,Suris2023ViperGPT,Wu2023VisualChatGPT,Yang2023MMReact,Liang2023TaskMatrix}.
However, no prior works have combined LLMs and different visual modules for the challenging text-to-image (T2I) generation task.
Our work proposes two novel interpretable/explainable visual programming (VP) frameworks combining LLMs and visual modules for T2I generation and evaluation. 

First, we introduce \textbf{\textboxmethod{}} (\cref{sec:model}), a new step-by-step T2I generation framework that decomposes the T2I task into three steps (object/count generation, layout generation, and image generation), where each step is implemented as a module and executed in an interpretable generation program,
as shown in \cref{fig:teaser} (a).
We employ Vicuna~\cite{vicuna2023}, a powerful publicly available LLM, to handle the first two steps (object/count generation and layout generation),
by finetuning it on text-layout pairs from multiple datasets~\cite{Plummer2017Flickr30kEntities,Lin2014COCO,Cho2023DallEval}, resulting in improved layout control for T2I generation.
For the last image generation step,
we use off-the-shelf layout-to-image generation models, \eg{} \gligen{}~\cite{Li2023GLIGEN}.
Our generation framework provides more interpretable spatial control than the widely used end-to-end T2I models~\cite{Rombach2021HighResolutionIS,Ramesh2021DALLE,Ramesh2022UnCLIP,Saharia2022Imagen}.
Furthermore, \textboxmethod{} leverages the world knowledge of pretrained LMs to understand unseen objects, overcoming the limitation of previous layout-guided T2I works that can only handle predefined object classes~\cite{Hong2018,Tan2019Text2Scene,Li2019f,Liang2022}.
In our analysis, we find that our \textboxmethod{} (\vicuna+\gligen{}) could generate images more accurately following the text description
(especially about object counts, spatial relations, and object sizes) than strong T2I generation baselines, such as \sd{}~\cite{Rombach2021HighResolutionIS} (see \cref{sec:eval_results_skill}).

Next, we introduce \textbf{\eval{}} (\cref{sec:vis_prog_evaluation}), a new interpretable/explainable T2I evaluation framework
based on evaluation programs that invoke diverse visual modules to evaluate different T2I skills~\cite{Cho2023DallEval} (see \cref{sec:evaluated_skills}),
and also provide visual+textual explanations of the evaluation results.
Previous T2I evaluation works have focused on measuring visual quality~\cite{Salimans2016a,Heusel2017}
and image-text alignment with a single visual module
based on text retrieval~\cite{Xu2018AttnGAN}, CLIP cosine similarity~\cite{Radford2021CLIP,Huang2021}, captioning~\cite{Hong2018}, object detection~\cite{Hinz2020,Cho2023DallEval,Gokhale2022VISOR}, and visual question answering (VQA)~\cite{hu2023tifa,yarom2023you}.
However, we cannot interpret the reasoning behind the scoring; \ie{} why CLIP assigns a higher score to an image-text pair than another.
In addition, these modules are good at measuring some skills, but not reliable in other skills
(\eg{} VQA/CLIP models are not good at counting objects or accurately parsing text rendered in images).
In \eval{}, we break the evaluation process into a mixture of visual evaluation modules, resulting in an interpretable evaluation program. The evaluation modules are experts in different skills and provide visual+textual (\ie{} multimodal) result/error explanations, as shown in \cref{fig:teaser} (b). We evaluate both multimodal LM and diffusion-based T2I models with two types of prompts: (1) skill-based prompts (\cref{sec:skill_based_eval}) and (2) open-ended prompts (\cref{sec:eval_prog_lm_eval}). Skill-based prompts evaluate a single skill per image, whereas open-ended prompts evaluate multiple skills per image. We adapt
a large language model
(GPT-3.5-Turbo~\cite{chatgpt}) to dynamically generate an evaluation program for each open-ended prompt, without requiring finetuning on expensive evaluation program annotations.

In our skill-based prompt experiments,
while all T2I models faced challenges in the \countskill{}, \spatialskill{}, \scaleskill{}, and \textskill{} skills, our \textboxmethod{} shows higher scores in the \countskill{}, \spatialskill{}, and \scaleskill{} skills, demonstrating its strong layout control.
We also show a fine-grained sub-split analysis on where our \textboxmethod{} gives an improvement over the
baseline T2I models.
In our open-ended prompt experiments, our \textboxmethod{} achieves competitive results to the T2I baselines.
Many of these open-ended prompts tend to focus more on objects and attributes, which, as shown by the skill-based evaluation, T2I models perform quite well in; however, when precise layouts and spatial relationships are needed, our \textboxmethod{} framework performs better (\cref{sec:eval_results_open}). In our error analysis, we also find that \gligen{} sometimes fails to properly generate images even when \vicuna{} generates the correct layouts, indicating that as better layout-to-image models become available, our \textboxmethod{} framework will achieve better results (\cref{sec:eval_results_open}). 
We also provide an analysis of the generated evaluation programs from \eval{}, which proved to be highly accurate and comprehensive in covering elements from the prompts (\cref{sec:human_correlation}).
For both types of prompts, our \eval{} method shows a higher correlation to human evaluation than existing single model-based evaluation methods (\cref{sec:human_correlation}).

Our contributions can be summarized as follows:
(1) \textbf{\textboxmethod{}} (\cref{sec:model}),
a new step-by-step T2I generation framework that decomposes the T2I task into three steps (object/count generation, layout generation, and image generation), where each step is implemented as a module and executed in an interpretable generation program;
(2)
\textbf{\eval{}} (\cref{sec:vis_prog_evaluation}), a new interpretable/explainable T2I evaluation framework
based on evaluation programs that execute diverse visual modules to evaluate different T2I skills and provide visual+textual explanations of the evaluation results;
(3) comprehensive analysis of different T2I models, which demonstrates the strong layout control of \textboxmethod{} and high human correlation of \eval{} (\cref{sec:experiments}).
We will release T2I model-generated images, \eval{} programs, and a public LM (finetuned for evaluation program generation using ChatGPT outputs).
We hope our research fosters future work on interpretable/explainable generation and evaluation for T2I tasks.

\section{Related Works}
\label{sec:related_works}

\noindent\textbf{Text-to-image generation models.} 
In the T2I generation task, models generate images from text.
Early deep learning models used the Generative Adversarial Networks (GAN)~\cite{Goodfellow2014} framework for this task \cite{Mansimov2016,Reed2016GAWWN,Reed2016GAN-INT-CLS,Xu2018AttnGAN}.
More recently, multimodal language models~\cite{Cho2020XLXMERT,Ramesh2021DALLE} and diffusion models~\cite{Sohl-Dickstein2015,Ho2020,Rombach2021HighResolutionIS,Nichol2022} have gained popularity.
Recent advances in multimodal language models such as Parti~\cite{Yu2022Parti} and MUSE~\cite{Chang2023Muse}, and diffusion models like \sd{}~\cite{Rombach2021HighResolutionIS}, UnCLIP~\cite{Ramesh2022UnCLIP}, and Imagen~\cite{Saharia2022Imagen}, have demonstrated a high level of photorealism in zero-shot image generation.

\par
\noindent\textbf{Bridging text-to-image generation with layouts.}
One line of research decomposes the T2I generation task into two stages: text-to-layout generation and layout-to-image generation~\cite{Hong2018,Tan2019Text2Scene,Li2019f,Liang2022}.
However, the previous approaches focus on a set of predefined object classes by training a new layout predictor module from scratch and therefore cannot place new objects unseen during training.
In contrast, our \textboxmethod{} uses an LM to handle layout generation by generating objects/counts/positions in text,
allowing flexible adaptation of pretrained LMs that can understand diverse region descriptions.

\par
\noindent\textbf{Language models with visual modules.}
Although large language models (LLMs) have shown a broad range of commonsense knowledge,
most of them are trained only on text corpus and cannot understand image inputs to tackle vision-and-language (VL) tasks.
Thus, recent works explore tackling VL tasks by solving sub-tasks with external visual modules and combining their outputs to obtain the final response~\cite{Yang2022PICa,Hu2022PromptCap,Gupta2023VisProg,Suris2023ViperGPT,Wu2023VisualChatGPT,Yang2023MMReact,Liang2023TaskMatrix}.
The visual sub-tasks include describing images as text,
finding image regions relevant to the text,
editing images with text guidance,
and obtaining answers from a VQA model.
However, existing work focuses on converting visual inputs into text format so that LLMs can understand them.
Our work is the first work using visual programming for interpretable and explainable T2I generation and evaluation.

\par
\noindent\textbf{Evaluation of text-to-image generation models.}
The text-to-image community has commonly used two types of automated evaluation metrics: visual quality and image-text alignment.
For visual quality, Inception Score (IS)~\cite{Salimans2016a} and Fréchet Inception Distance (FID)~\cite{Heusel2017} have been widely used.
For image-text alignment, previous work used a single model to calculate an alignment score for image-text pair,
based on text retrieval~\cite{Xu2018AttnGAN}, cosine similarity~\cite{Huang2021}, captioning~\cite{Hong2018}, object detection~\cite{Hinz2020,Cho2023DallEval,Gokhale2022VISOR}, and visual question answering (VQA)~\cite{hu2023tifa,yarom2023you}.
In this work,
we propose the first T2I evaluation framework \eval{},
based on interpretable and explainable evaluation programs which execute a diverse set of visual modules
(\eg{} object detection, OCR, depth estimation, object counting).
Our \eval{} evaluation programs provide visual+textual explanations of the evaluation result and demonstrate a high correlation with human judgments.

\begin{figure}
  \centering
  \includegraphics[width=.9\linewidth]{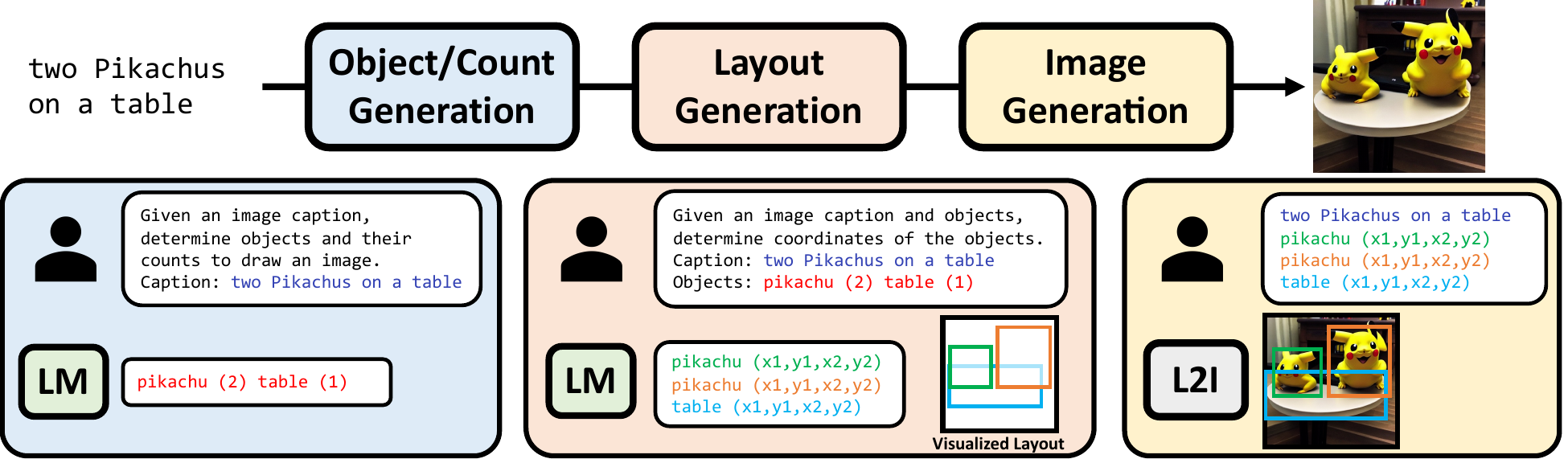}
  \caption{
  Interpretable step-by-step text-to-image generation with \textboxmethod{}. \textboxmethod{} decomposes T2I generation into three steps: (1) object/count generation, (2) layout generation, and (3) image generation,
  and executes the three modules step-by-step.
  }
\label{fig:vpgen} 
\vspace{-10pt}
\end{figure}

\section{\textboxmethod{}: Visual Programming for Step-by-Step Text-to-Image Generation}
\label{sec:model}
We propose \textboxmethod{}, a novel visual programming framework for interpretable step-by-step text-to-image (T2I) generation.
As illustrated in \cref{fig:vpgen},
we decompose the text-to-image generation task into three steps:
(1) object/count generation, (2) layout generation, and (3) image generation.
In contrast to previous T2I generation works that use an intermediate layout prediction module~\cite{Hong2018,Tan2019Text2Scene,Li2019f,Liang2022},
\textboxmethod{} represents all layouts (object description, object counts, and bounding boxes) in text,
and employs
an LM to handle the first two steps:
(1) object/count generation and (2) layout generation.
This makes it easy to adapt the knowledge of pretrained LMs and enables generating layouts of objects that are unseen during text-to-layout training (\eg{} `pikachu').
For layout representation, we choose the bounding box format because of its efficiency; bounding boxes generally require fewer tokens than other formats.\footnote{For example, a xyxy-format bounding box can be represented with 4 tokens, while a 64x64 segmentation map requires 4096 tokens.}

\par
\noindent\textbf{Two-step layout generation with LM.}
\cref{fig:vpgen} illustrates how our LM generates layouts in two steps:
(1) object/count generation and
(2) layout generation.
For the first step,
we represent the scene by enumerating objects and their counts, such as \texttt{``obj1 (\# of obj1) obj2 (\# of obj2)''}.
For the second step, following previous LM-based object detection works~\cite{Chen2022Pix2Seq,Yang2022UniTAB},
we normalize `xyxy' format bounding box coordinates into $[0, 1]$ and quantize them into 100 bins; a single object is represented as \texttt{``obj (xmin,ymin,xmax,ymax)''}, where each coordinate is within $\{0, \cdots, 99\}$.

\par
\noindent\textbf{Training layout-aware LM.}
To obtain the layout-aware LM,
we use Vicuna 13B~\cite{vicuna2023},
a public
state-of-the-art
language model finetuned from LLaMA~\cite{Touvron2023LLaMA}.
We use parameter-efficient finetuning with LoRA~\cite{Hu2022LoRA} to preserve the original knowledge of the LM and save memory during training and inference.
We collect text-layout pair annotations from training sets of three public datasets:
Flickr30K entities~\cite{Plummer2017Flickr30kEntities}, MS COCO instances 2014~\cite{Lin2014COCO}, and PaintSkills~\cite{Cho2023DallEval}, totaling 1.2M examples. See appendix for more training details.

\par
\noindent\textbf{Layout-to-Image Generation.}
We use a recent layout-to-image generation model \gligen{}~\cite{Li2023GLIGEN} for the final step - image generation.
The layout-to-image model takes a list of regions (bounding boxes and text descriptions) as well as the original text prompt to generate an image.

\section{\eval{}: Visual Programming for Explainable Evaluation of Text-to-Image Generation}
\label{sec:vis_prog_evaluation}

\begin{figure}
  \centering
  \includegraphics[width=.95\linewidth]{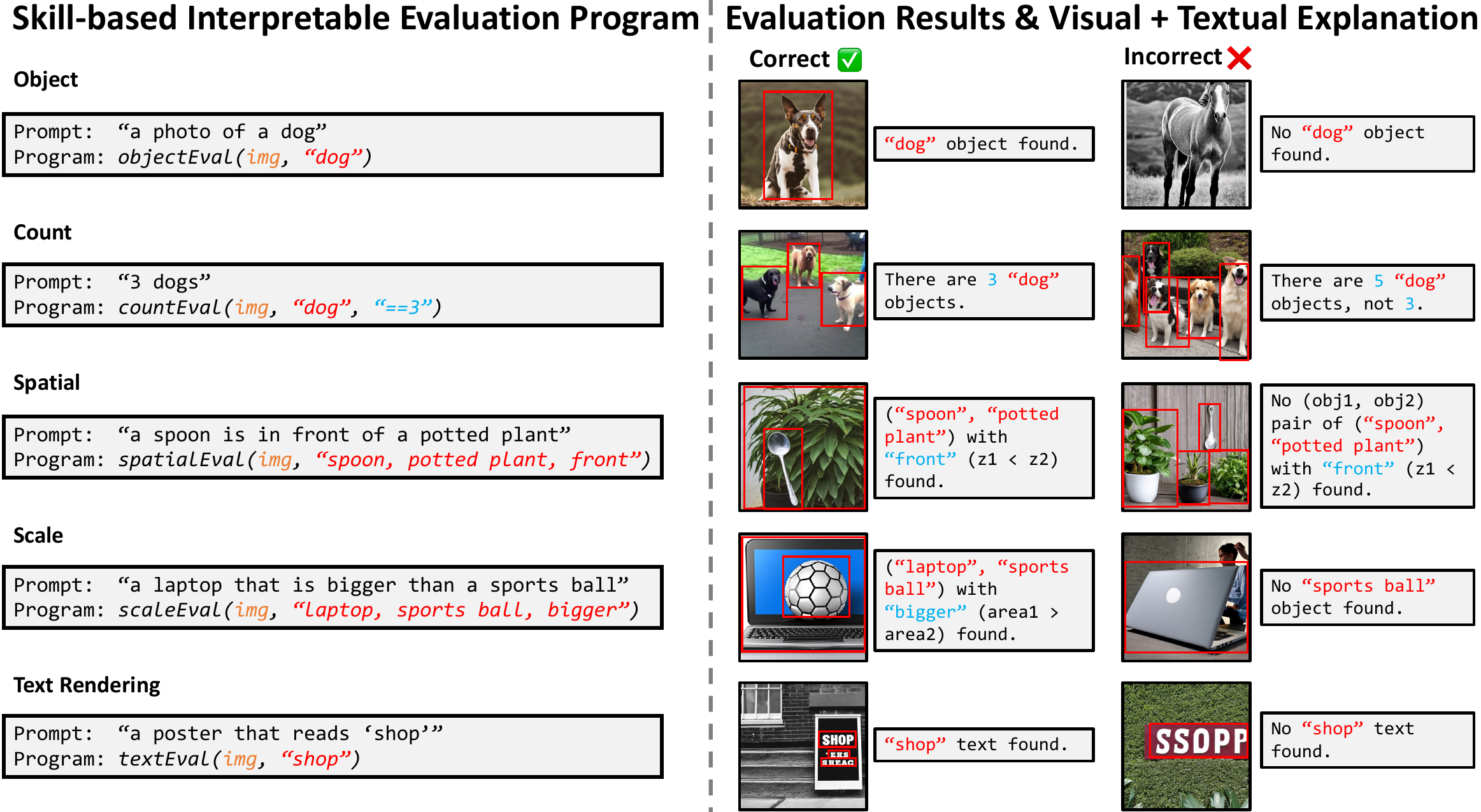}
  \caption{
  Illustration of skill-based evaluations in \eval{} (\cref{sec:skill_based_eval}).
  \textbf{Left}: Given text prompts that require different image generation skills,
  our interpretable evaluation programs evaluate images by executing relevant visual modules.
  \textbf{Right}:
  Our evaluation programs output binary scores and provide visual (bounding boxes of detected objects/text) + textual explanations of the evaluation results.
  }
\label{fig:vpeval_skill} 
\vspace{-10pt}
\end{figure}

\eval{} is a novel interpretable/explainable evaluation framework for T2I generation models, based on visual programming.
Unlike existing T2I evaluation methods that compute image-text alignment scores with an end-to-end model,
our evaluation 
provides an interpretable program and visual+textual explanations for the evaluation results,
as shown in \cref{fig:vpeval_skill,fig:open_eval_program_generation}.
We propose two types of evaluation prompts: (1) skill-based evaluation and (2) open-ended evaluation.
In skill-based evaluation, 
we define five image generation skills and use a set of skill-specific prompts and evaluation programs,
as illustrated in \cref{fig:vpeval_skill}.
In open-ended evaluation,
we use a diverse set of prompts that require multiple image generation skills.
We adopt a language model to dynamically generate an evaluation program for each text prompt, as shown in \cref{fig:open_eval_program_generation}.
In the following, we describe
evaluation skills (\cref{sec:evaluated_skills}),
visual evaluation modules (\cref{sec:module_definitions}),
skill-based evaluation with visual programs (\cref{sec:skill_based_eval}),
and open-ended evaluation with visual program generator LM (\cref{sec:eval_prog_lm_eval}).

\subsection{Evaluation Skills}
\label{sec:evaluated_skills}

Inspired by skill-based T2I analysis of PaintSkills~\cite{Cho2023DallEval},
our \eval{} measures five image generation skills:
\objectskill{}, \countskill{}, \spatialskill{}, \scaleskill{}, and \textskill{}.
Powered by evaluation programs with expert visual modules,
our \eval{} supports
zero-shot evaluation of images (no finetuning of T2I models is required),
detecting regions with free-form text prompts,
new 3D spatial relations (front, behind),
and new scale comparison and text rendering skills,
which were not supported in PaintSkills~\cite{Cho2023DallEval}.
In \cref{fig:vpeval_skill}, we illustrate the evaluation process for each skill.

\par
\noindent\textbf{\objectskill{}.}
Given a prompt with an object (\eg{} ``a photo of a dog''), a T2I model should generate an image with that object present.

\par
\noindent\textbf{\countskill{}.}
Given a prompt with a certain number of an object (\eg{} ``3 dogs''), a T2I model should generate an image containing the specified number of objects.

\par
\noindent\textbf{\spatialskill{}.}
Given a prompt with two objects and a spatial relationship between them (\eg{} ``a spoon is in front of a potted plant''), a T2I model should generate an image that contains both objects with the correct spatial relations.

\par
\noindent\textbf{\scaleskill{}.}
Given a prompt with two objects and a relative scale between them (\eg{} ``a laptop that is bigger than a sports ball''), a T2I model should generate an image that contains both objects, and each object should be of the correct relative scale.

\par
\noindent\textbf{\textskill{}.}
Given a prompt with a certain text to display (\eg{} ``a poster that reads \texttt{`shop'}''), a T2I model should generate an image that properly renders the text.

\begin{figure}
  \centering
  \includegraphics[width=.9\linewidth]{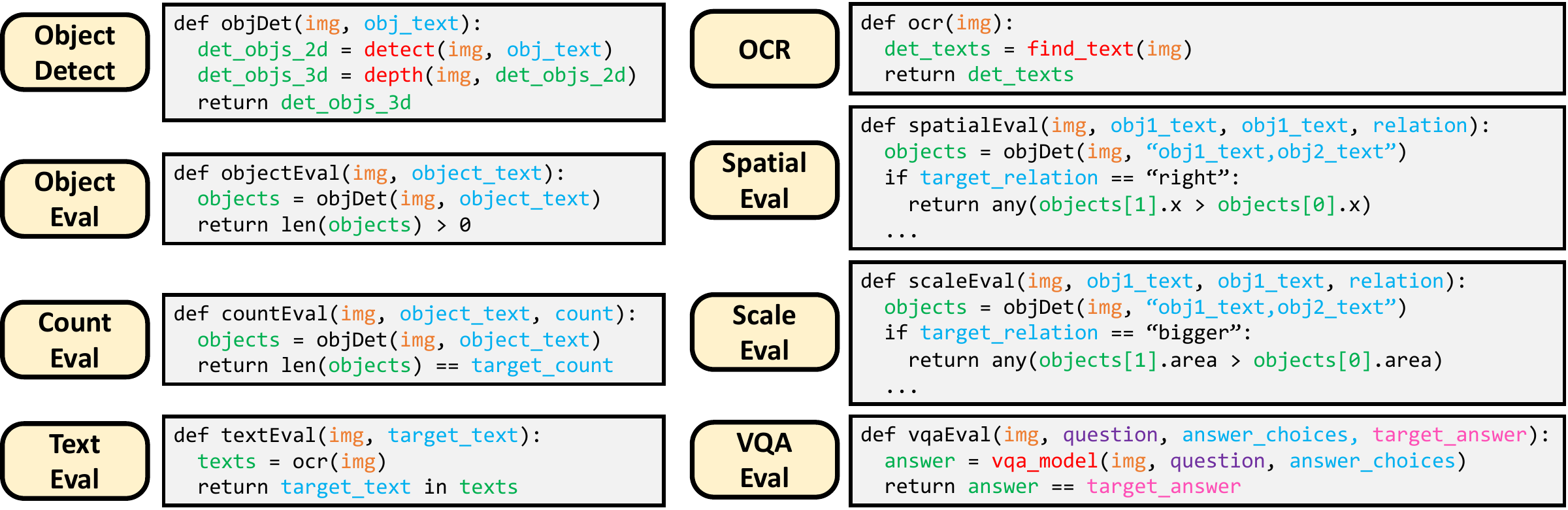}
  \caption{Python pseudocode implementation of visual modules used in \eval{}.
  \vspace{-10pt}
  }
\label{fig:vis_modules} 
\end{figure}

\subsection{Visual Evaluation Modules}
\label{sec:module_definitions}

To measure the skills described above in \cref{sec:evaluated_skills},
we use eight expert
visual evaluation modules specialized for different tasks.
The modules provide visual+textual explanations with their score.
Visual explanations are generated by rendering bounding boxes on the images and textual explanations are generated through text templates, as shown in \cref{fig:vpeval_skill} and \ref{fig:open_eval_program_generation}.
We provide the Python pseudocode implementation for each module
in \cref{fig:vis_modules}.

\par
\noindent\textbf{Module definitions.}
\textbf{objDet} detects objects in an image based on a referring expression text and returns them (+ their 2D bounding boxes and depth), using Grounding DINO~\cite{ShilongLiu2023GroundingDino} and DPT~\cite{Ranftl2021VisionTF}.
\textbf{ocr} detects all text in an image and returns them (+ their 2D bounding boxes), using EasyOCR~\cite{EasyOCR2023}.
\textbf{vqa} answers a multiple-choice question using BLIP-2 (Flan-T5 XL)~\cite{Li2023BLIP2}. It can handle phrases that cannot be covered only with \texttt{objDet} or \texttt{ocr} (\eg{} pose estimation, action recognition, object attributes).
\textbf{objectEval} evaluates if an object is in an image using \texttt{objDet}.
\textbf{countEval} evaluates if an object occurs in an image a certain number of times using \texttt{objDet} and an equation (\eg{} ``==3'', ``<5'').
\textbf{spatialEval} evaluates if two objects have a certain spatial relationship with each other. For six relations (above, below, left, right, front, and behind), it compares bounding box/depth values using \texttt{objDet}.
For other relations, it uses \texttt{vqa}.
\textbf{scaleEval} evaluates if two objects have a certain scale relationship with each other using \texttt{objDet}. For three scale relations (smaller, bigger, and same) it compares bounding box areas by using \texttt{objDet}.
For other relations, it uses \texttt{vqa}.
\textbf{textEval} evaluates if a given text is present in an image using \texttt{ocr}.

\subsection{Skill-based Evaluation with Visual Programs}
\label{sec:skill_based_eval}

For skill-based evaluation,
we create text prompts with various skill-specific templates that are used for image generation and evaluation with our programs. See appendix for the details of prompt creation.
In \cref{fig:vpeval_skill}, we illustrate our skill-based evaluation in \eval{}.
Given text prompts that require different image generation skills,
our evaluation programs measure image-text alignment scores by calling the relevant visual modules.
Unlike existing T2I evaluation methods,
our evaluation programs provide visual+textual explanations of the evaluation results.

\subsection{Open-ended Evaluation with Visual Program Generator LM}
\label{sec:eval_prog_lm_eval}

\begin{figure}
  \centering
  \includegraphics[width=.95\linewidth]{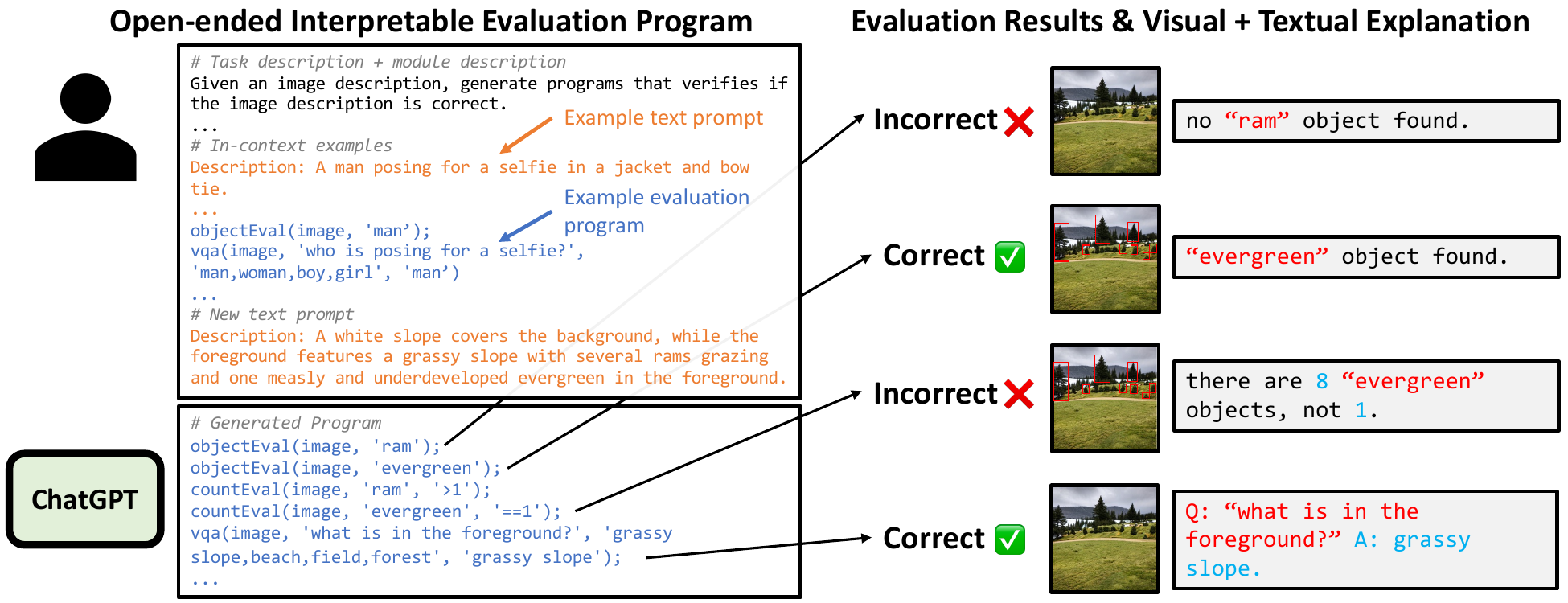}
  \caption{
  Evaluation of open-ended prompts in \eval{} (\cref{sec:eval_prog_lm_eval}).
  \textbf{Left}: We generate evaluation programs with ChatGPT via in-context learning.
  \textbf{Right}: Our evaluation programs consist of multiple modules evaluating different elements from a text prompt. We output the final evaluation score by averaging the outputs of each evaluation module.
  }
\label{fig:open_eval_program_generation} 
\end{figure}

Although our evaluation with skill-specific prompts covers five important and diverse image generation skills,
user-written prompts can sometimes be even more complex and need multiple evaluation criteria (\eg{} a mix of our skills in \cref{sec:evaluated_skills} and other skills like attribute detection).
For example, evaluating images generated with the prompt `A woman dressed for a wedding is showing a watermelon slice to a woman on a scooter.'
involves multiple aspects, such as
two women (count skill), `a woman on a scooter' (spatial skill), `dressed for a wedding' (attribute detection skill), \etc{}.
To handle such open-ended prompts,
we extend the \eval{} setup with evaluation programs that can use many visual modules together (whereas single-skill prompts can be evaluated with a program of 1-2 modules).
We generate open-ended prompt evaluation programs with an LLM, then the evaluation programs output the average score and the visual+textual explanations from their visual modules. The program generation involves choosing which prompt elements to evaluate and which modules to evaluate those elements (see \cref{fig:open_eval_program_generation}).

\par
\noindent\textbf{Open-ended prompts.}
For open-ended evaluation,
we use 160 prompts of TIFA v1.0 human judgment dataset~\cite{hu2023tifa} (we refer to these prompts as `\tifaprompts{}').
The dataset consists of (1) text prompts from COCO~\cite{Lin2014COCO}, PaintSkills~\cite{Cho2023DallEval}, DrawBench~\cite{Saharia2022Imagen}, and Partiprompts~\cite{Yu2022Parti}, (2) images generated by five baseline models (minDALL-E~\cite{kakaobrain2021minDALL-E}, VQ-Diffusion~\cite{Gu2022VQdiffusion}, Stable Diffusion v1.1/v1.5/v2.1~\cite{Rombach2021HighResolutionIS}), and (3) human judgment scores on the images (on 1-5 Likert scale).

\par
\noindent\textbf{Generating evaluation programs via in-context learning.}
As annotation of evaluation programs with open-ended prompts can be expensive, we use ChatGPT (GPT-3.5-Turbo)~\cite{chatgpt} to generate evaluation programs via in-context learning.
To guide ChatGPT, we adapt the 12 in-context prompts from TIFA~\cite{hu2023tifa}.
We show ChatGPT the
list of visual modules and example text prompts and programs,
then ask the model to generate a program given a new prompt, as illustrated in \cref{fig:open_eval_program_generation}.
For reproducible and accessible evaluation, we release the evaluation programs so that \eval{} users do not have to generate the programs themselves.
We will also release a public LM (finetuned for evaluation program generation using ChatGPT outputs) that can run on local machines.

\section{Experiments and Results}
\label{sec:experiments}

In this section,
we introduce the baseline T2I models we evaluate (\cref{sec:evaluated_models}),
compare different T2I models with skill-based (\cref{sec:eval_results_skill}) and open-ended prompts (\cref{sec:eval_results_open}),
and show the human-correlation of \eval{} (\cref{sec:human_correlation}).

\subsection{Evaluated Models}
\label{sec:evaluated_models}

We evaluate our \textboxmethod{} (\vicuna+\gligen{}) and five popular and publicly available T2I models, covering both diffusion models (\sd{} v1.4/v2.1~\cite{Rombach2021HighResolutionIS}
and \karlo{}~\cite{kakaobrain2022karlo-v1-alpha}), and multimodal autoregressive language models (\mindalle{}~\cite{kakaobrain2021minDALL-E} and \dallemega{}~\cite{dallemega2021}).
\sdcompvis{} is the most comparable baseline to \textboxmethod{}{},
because the \gligen{}~\cite{Li2023GLIGEN} in \textboxmethod{} uses frozen \sdcompvis{} with a few newly inserted adapter parameters for spatial control.

\subsection{Evaluation on Skill-based Prompts}
\label{sec:eval_results_skill}

\begin{table}
  \caption{
  \eval{} scores of T2I generation models on skill-based prompts (see \cref{sec:eval_results_skill} for analysis).
  \textit{F30: Flickr30K Entities, C: COCO, P: PaintSkills}.
  }
  \label{tab:skill_eval_results}
  \centering
  \resizebox{0.9\columnwidth}{!} {
      \begin{tabular}{l c c c c c c c}
        \toprule
        \multirow{2}{*}{Model} & \multicolumn{6}{c}{\eval{} Skill Score (\%) $\uparrow$} \\
        \cmidrule(r){2-7}
        & \objectskill{} & \countskill{} & \spatialskill{} & \scaleskill{} & \textskill{} & Average \\
        \midrule
        \sdcompvis{} & \textbf{97.3} & 47.4 & 22.9 & 11.9 & \textbf{8.9} & 37.7\\
        \sdstabai{} & 96.5 & 53.9 & 31.3 & 14.3 & 6.9 & 40.6\\
        \karlo{} & 95.0 & 59.5 & 24.0 & 16.4 & \textbf{8.9} & 40.8\\
        \mindalle{} & 79.8 & 29.3 & 7.0 & 6.2 & 0.0 & 24.4\\
        \dallemega{} & 94.0 & 45.6 & 17.0 & 8.5 & 0.0 & 33.0\\
        \midrule

        \textboxmethod{} (F30) & 96.8 & 55.0 & 39.0 & 23.3 & 5.2 & 43.9 \\
        \textboxmethod{} (F30+C+P) & 96.8 & \textbf{72.2} & \textbf{56.1} & \textbf{26.3} & 3.7 & \textbf{51.0} \\
        \bottomrule
      \end{tabular}
  }
  \vspace{-10pt}
\end{table}

\par
\noindent\textbf{Diffusion models outperform multimodal LMs.}
In \Cref{tab:skill_eval_results}, we show the \eval{} skill accuracies for each model.
The diffusion models (\sd{}, \karlo{}, and our \textboxmethod{}) show higher overall accuracy than the multimodal LMs (\mindalle{} and \dallemega{}).

\par
\noindent\textbf{Count/Spatial/Scale/Text Rendering skills are challenging.}
Overall, the five baseline T2I models achieve high scores (above 93\% except for \mindalle{}) in \objectskill{} skill; \ie{} they are good at generating a high-quality single object. 
However, the models score low accuracies in the other skills, indicating that these skills (\countskill{}, \spatialskill{}, \scaleskill{}, and \textskill{}) are still challenging with recent T2I models.

\par
\noindent\textbf{Step-by-step generation improves challenging skills.}
The bottom row of \Cref{tab:skill_eval_results} shows that our \textboxmethod{} achieves high accuracy in \objectskill{} skill and strongly outperforms other baselines in \countskill{} (+12.7\% than Karlo), \spatialskill{} (+24.8\% than \sdstabai{}) and \scaleskill{} (+9.9\%  than Karlo) skills.
This result demonstrates that our step-by-step generation method is effective in aligning image layout with text prompts, while also still having an interpretable generation program.
All models achieve low scores on the \textskill{} skill, including our \textboxmethod{}. A potential reason why our \textboxmethod{} model does not improve \textskill{} score might be because the text-to-layout
training datasets of \textboxmethod{} (Flickr30K Entities~\cite{Plummer2017Flickr30kEntities}, COCO~\cite{Lin2014COCO}, and PaintSkills \cite{Cho2023DallEval}) contain very few images/captions about text rendering, opening room for future improvements (\eg{} finetuning on a dataset with more images that focus on text rendering and including a text rendering module as part of our generation framework).

\begin{figure}[h]
  \includegraphics[width=.9\linewidth]{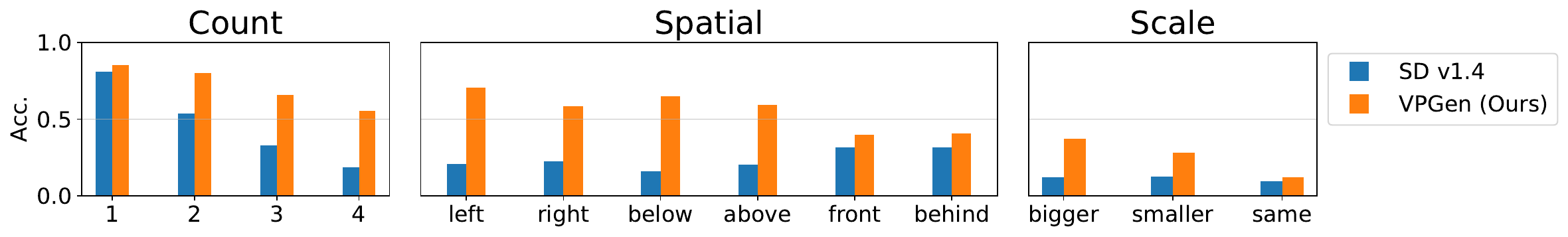}
  \centering
  \caption{
  \eval{} score comparison of our \textboxmethod{} (\vicuna+\gligen{}) and \sdcompvis{} on
  fine-grained splits in \countskill{}/\spatialskill{}/\scaleskill{} skills. 
  }
\label{fig:fine_grained_skills} 
\vspace{-10pt}
\end{figure}

\noindent\textbf{Fine-grained analysis in \countskill{}/\spatialskill{}/\scaleskill{} skills.}
To better understand the high performance of our \textboxmethod{} in \countskill{}, \spatialskill{}, and \scaleskill{} skills,
we perform a detailed analysis of these skills with fine-grained splits.
In \cref{fig:fine_grained_skills}, we compare our \textboxmethod{} (\vicuna+\gligen{}) and its closest baseline \sd{} v1.4 on the three skills.
\textbf{Overall}: \textboxmethod{} achieves better performance than \sd{} on every split on all three skills.
\textbf{\countskill{}}: 
While both models show difficulties with counting larger numbers,
our \textboxmethod{} model achieves better performance with 50+ accuracy on all four numbers.
\textbf{\spatialskill{}}:
Our \textboxmethod{} model performs better on all six spatial relations.
\textboxmethod{} shows higher accuracy on 
2D relations (left/right/below/above) than on 3D depth relations (front/behind), while the \sd{} is better at 3D relations than 2D relations.
\textbf{\scaleskill{}}:
\textboxmethod{} generates more accurate layouts with two objects of different sizes (bigger/smaller), than layouts with two objects of similar sizes (same).

\begin{wraptable}[14]{r}{55mm}
\centering
\vspace{-38pt}
  \caption{
  \eval{} scores on open-ended prompts (see \cref{sec:eval_results_open}).
  \gligen{} in \textboxmethod{} has \sdcompvis{} backbone.
  \textit{F30: Flickr30K Entities, C: COCO, P: PaintSkills}.
  }
  \label{tab:open_eval_results}
\vspace{6pt}
  \resizebox{0.4\columnwidth}{!} {
      \begin{tabular}{l c}
        \toprule
        \multirow{1}{*}{Model} & \multicolumn{1}{c}{Score (\%) $\uparrow$}\\
        \midrule
        \sdcompvis{} & 70.6 \\
        \sdstabai{}  & 72.0 \\
        \karlo{}     & 70.0 \\
        \mindalle{}  & 47.5 \\
        \dallemega{} & 67.2 \\
        \midrule
        \textboxmethod{} (F30) & 71.0  \\
        \textboxmethod{} (F30+C+P) & 68.3  \\
        \bottomrule
      \end{tabular}
  }
\end{wraptable}

\subsection{Evaluation on Open-ended Prompts}
\label{sec:eval_results_open}

\Cref{tab:open_eval_results} shows the \eval{} score on the open-ended \tifaprompts{} prompts. We calculate the score by averaging accuracy from the modules (see \cref{fig:open_eval_program_generation} for an example).
The overall score trend of the baseline models (diffusion models > multimodal LMs) is consistent with the skill-based prompts.
Unlike the trend in skill-based evaluation (\cref{sec:eval_results_skill}),
our \textboxmethod{} (\vicuna{}+\gligen{}) methods achieve similar performance to the \sd{} baseline, while also providing interpretable generation steps.
We perform an analysis of the open-ended prompts used for evaluation. In these prompts, object descriptions and attributes are the dominating prompt element (86.4\% of elements), whereas descriptions of spatial layouts, only account for 13.6\% of elements. See appendix for more details.
For prompts from PaintSkills~\cite{Cho2023DallEval} where spatial layouts are more important, we find that our \textboxmethod{} scores much higher than \sdcompvis{} (71.0 \vs{} 63.5) and \sdstabai{} (71.0 \vs{} 68.4). Note that \gligen{} used in \textboxmethod{} is based on \sdcompvis{}, making it the closest baseline, and hence future versions of \gligen{} based on \sdstabai{} or other stronger layout-to-image models will improve our results further.
Also, we find that \gligen{} sometimes fails to properly generate images even when \textboxmethod{} generates the correct layouts (see following paragraph).

\noindent\textbf{\textboxmethod{} sources of error: layout/image generation.}
Since our \textboxmethod{} separates layout generation and image generation,
we study the errors caused by each step.
For this, we manually analyze the images generated from \tifaprompts{} prompts,
and label
(1) whether the generated layouts align with the text prompts
and (2) whether the final images align with the text prompts/layouts.
\gligen{} sometimes fails to properly generate images even when \vicuna{} 13B generates the correct layouts, indicating that when better layout-to-image models become available, our \textboxmethod{} framework will achieve higher results.
We include more details in the appendix.

\begin{figure}[t]
  \includegraphics[width=.9\linewidth]{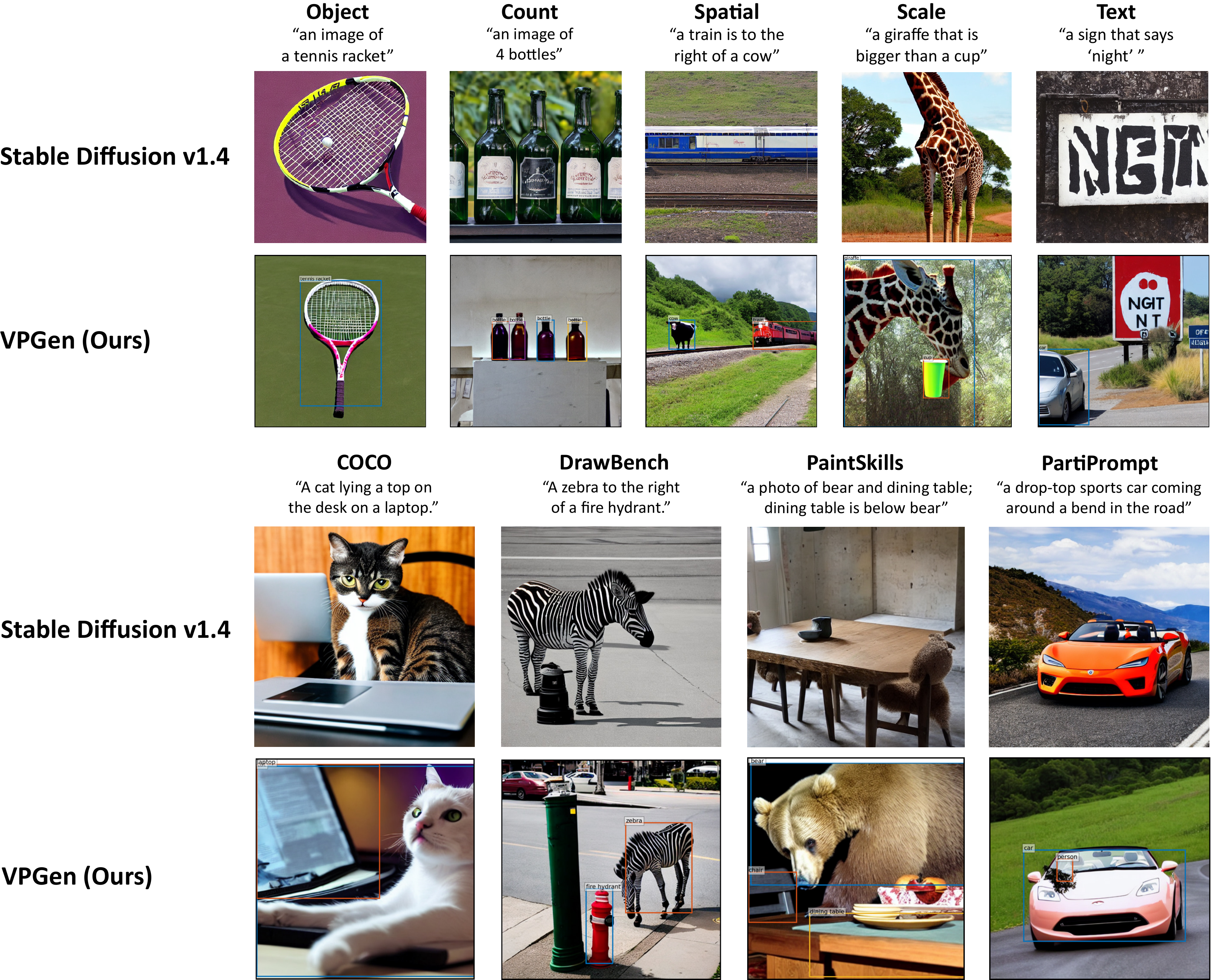}
  \centering
  \caption{
  Images generated by \sdcompvis{} and our \textboxmethod{} (\vicuna{} 13B+\gligen{}) for skill-based (top) and open-ended (bottom) prompts.
  }
\label{fig:main_qualitative_examples} 
\vspace{-10pt}
\end{figure}

\noindent\textbf{Qualitative Examples.}
In \cref{fig:main_qualitative_examples}, we provide example images generated by our \textboxmethod{} and \sdcompvis{} (the closest baseline) with various skill-based (top) and open-ended (bottom) prompts. See appendix for more qualitative examples with skill-based/open-ended prompts, prompts with unseen objects, counting $\ge4$ objects, and error analysis.

\subsection{Human Evaluation of \eval{}}
\label{sec:human_correlation}

To measure alignments with human judgments and our \eval{},
we compare the human-correlations of \eval{} and other metrics
on both skill-based prompts and open-ended prompts.

\begin{table}
  \caption{
    Human correlation study of skill-based evaluation.
  We measure Spearman's $\rho$ correlation between human judgment and different automated metrics on the skill-based prompts (\cref{sec:skill_based_eval}).
  \eval{}$^\dagger$: using BLIP-2 VQA for \texttt{objectEval/spatialEval/scaleEval} modules. 
  }
  \label{tab:human_corelations_skill}
  \centering
  \resizebox{0.9\columnwidth}{!}{
      \begin{tabular}{l ccccc c}
        \toprule
        \multirow{2}{*}{Eval Metric} & \multicolumn{6}{c}{Human-metric correlation (Spearman's $\rho$) $\uparrow$} \\
        \cmidrule(r){2-7}
        & Object & Count & Spatial & Scale & Text & Overall \\
        \midrule
        CLIP Cosine similarity (ViT-B/32) & 35.2 & 38.6 & 35.4 & 13.7 & 40.0 & 20.4 \\
        BLIP-2 Captioning - BLEU & 11.9 & 31.4 & 26.3 & 24.0 & 23.6 & -3.4 \\
        BLIP-2 Captioning - ROUGE & 15.7 & 26.5 & 28.0 & 12.2 & 28.3 & 11.9 \\
        BLIP-2 Captioning - METEOR & 33.7 & 20.7 & 40.5 & 25.1 & 26.6 & 29.3 \\
        BLIP-2 Captioning - SPICE & 56.1 & 20.9 & 40.6 & 27.3 & 18.6 & 28.1 \\
        BLIP-2 VQA & \textbf{63.7} & 63.1 & 38.9 & 26.1 & 31.3 & 65.0 \\
        \midrule
        \eval{} & 34.5 & \textbf{63.8} & 48.9 & 29.4 & \textbf{85.7} & 73.5 \\
        \eval{}$^\dagger$ & \textbf{63.7} & \textbf{63.8} & \textbf{51.2} & \textbf{29.5} & \textbf{85.7} & \textbf{79.0} \\
        \bottomrule
      \end{tabular}
  }
  \vspace{-10pt}
\end{table}

\vspace{-3pt}
\noindent\textbf{Human correlation of \eval{} on skill-based prompts.}
We ask two expert annotators to evaluate 20 images, from each of the five baseline models, for each of the five skills with binary scoring (total $20\times5\times5=500$ images).
The inter-annotator agreements were measured with
Cohen's $\kappa$~\cite{cohen1960coefficient} 
= 0.85
and
Krippendorff's $\alpha$~\cite{krippendorff_alpha}
= 0.85,
indicating `near-perfect' ($\kappa$ > 0.8 or $\alpha$ > 0.8) agreement~\cite{McHugh2012-qw,kappareliablityJulius,krippendorff_alpha}.
We compare our visual programs with captioning (with metrics BLEU~\cite{Papineni2002}, ROUGE~\cite{Lin2004}, METEOR~\cite{Banerjee2005}, and SPICE~\cite{Anderson2016}), VQA, and CLIP (ViT-B/32) based evaluation.
We use BLIP-2 Flan-T5 XL, the state-of-the-art public model for image captioning and VQA.

In \Cref{tab:human_corelations_skill},
our \eval{} shows a higher overall correlation with human judgments (66.6) than single module-based evaluations (CLIP, Captioning, and VQA).
Regarding per-skill correlations, \eval{} shows especially strong human correlations in \countskill{} and \textskill{}.
As the BLIP-2 VQA module also shows strong correlation on \objectskill{}/\spatialskill{}/\scaleskill{},
we also experiment with \eval{}$^\dagger$: using BLIP-2 VQA for \texttt{objectEval/spatialEval/scaleEval} modules (instead of Grounding DINO and DPT),
which increases human correlation scores.
Note that our object detection modules visually explain the evaluation results, as shown in \cref{fig:vpeval_skill},
and we can have an even higher correlation when we have access to stronger future object detection models.

\begin{wraptable}[17]{r}{45mm}
\centering
\vspace{-20pt}
\caption{Human correlation on open-ended evaluation
with Spearman's $\rho$.
}
\vspace{6pt}
\label{tab:human_correltion_tifa}

  \resizebox{0.3\columnwidth}{!} {
  \begin{tabular}{l c}
    \toprule
    Metrics & $\rho$ ($\uparrow$)  \\
    \midrule
     \textit{BLIP-2 Captioning} \\
     BLEU-4 & 18.3 \\
     ROUGE-L & 32.9  \\
     METEOR & 34.0 \\
     SPICE & 32.8 \\
     \midrule
     \textit{Cosine-similarity} \\
     CLIP (ViT-B/32) & 33.2 \\
     \midrule
     \textit{LM + VQA module} \\
     TIFA (BLIP-2) & 55.9 \\
     \midrule
     \textit{LM + multiple modules} \\
     \eval{} (Ours) & 56.9\\
     \eval{}$^\dagger$ (Ours) & \textbf{60.3}  \\
    \bottomrule
  \end{tabular}
  }
\end{wraptable}

\vspace{-3pt}
\noindent\textbf{Human correlation of \eval{} on open-ended prompts.}
We generate visual programs with our program generation LM on \tifaprompts{} prompts~\cite{hu2023tifa}.
The dataset consists of 800 images (160 prompts $\times$ 5 T2I models)
and human judgments (1-5 Likert scale) along with other automatic metrics (BLIP-2 captioning, CLIP cosine similarity, and TIFA with BLIP-2) on the images.
\Cref{tab:human_correltion_tifa} shows that
our \eval{} achieves a better human correlation with TIFA (BLIP-2), and our \eval{}$^\dagger$ version achieves an even higher correlation.
The results indicate that using various interpretable modules specialized in different skills complements each other and improves human correlation, while also providing visual+textual explanations.

\vspace{-3pt}
\noindent\textbf{Human analysis on the generated programs.}
Lastly, we also measure the faithfulness of the generated evaluation programs.
For this, we sample \tifaprompts{} prompts and analyze the evaluation programs by
(1) how well the modules cover elements in the prompt;
(2) how accurate the module outputs are when run.
We find that programs generated by our \eval{} have very high coverage over the prompt elements and high per-module accuracy compared to human judgment.
We include more details in the appendix.

\section{Conclusion}

We propose two novel visual programming frameworks for interpretable/explainable T2I generation and evaluation: \textboxmethod{} and \eval{}.
\textboxmethod{} is a step-by-step T2I generation framework that decomposes the T2I task into three steps (object/count generation, layout generation, and image generation), leveraging an LLM for the first two steps,
and layout-to-image generation models for the last image generation step.
\textboxmethod{} generates images
more accurately following the text descriptions (especially about object counts, spatial relations, and object sizes) than strong T2I baselines,
while still having an interpretable generation program.
\eval{} is a T2I evaluation framework that uses interpretable evaluation programs with diverse visual modules that are experts in different skills to measure various T2I skills and provide visual+textual explanations of evaluation results.
In our analysis,
\eval{} presents a higher correlation with human judgments than single model-based evaluations, on both skill-specific and open-ended prompts.
We hope our work encourages future progress on interpretable/explainable generation and evaluation for T2I models.
\par
\textbf{Limitations \& Broader Impacts.} See Appendix for limitations and broader impacts discussion.

\section*{Acknowledgement}

We thank the reviewers for their valuable comments and feedback.
This work was supported by ARO W911NF2110220, DARPA MCS N66001-19-2-4031, NSF-AI Engage Institute DRL211263, ONR N00014-23-1-2356, and DARPA ECOLE Program No. HR00112390060. The views, opinions, and/or findings contained in this article are those of the authors and not of the funding agency.

{
\small
\bibliographystyle{unsrt}
\bibliography{citations}
}

\newpage
\appendix

\section*{Appendix}

In this appendix,
we provide
qualitative examples (\cref{sec:qual_examples}),
additional \textboxmethod{} analysis (\cref{sec:gen_model_additional_experiments}),
\textboxmethod{} error analysis (\cref{sec:vpgen_error_analysis}),
\eval{} error analysis (\cref{sec:vpeval_error_analysis}),
\textboxmethod{}/\eval{} implementation details (\cref{sec:implemenation_details}),
TIFA prompt element analysis details (\cref{sec:tifa_prompt_element_analysis_details}),
limitations and broader impacts
(\cref{sec:limitation}),
and license information (\cref{sec:license}).

\section{Qualitative Examples}
\label{sec:qual_examples}
\begin{figure}[h]
  \centering
  \includegraphics[width=.95\linewidth]{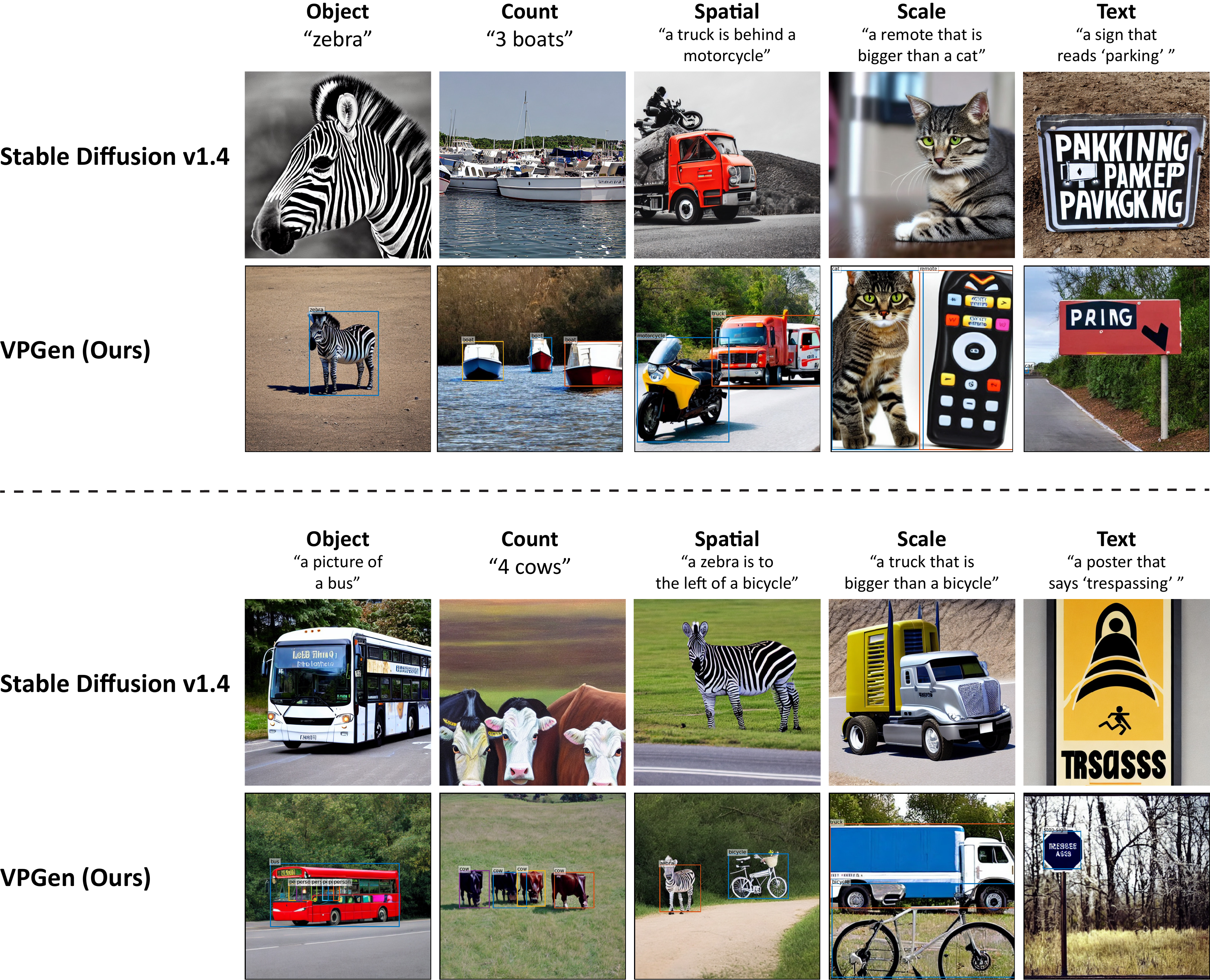}
  \caption{
    Images generated by \sdcompvis{} and our \textboxmethod{} (\vicuna{} 13B+\gligen{}) for our skill-based prompts.
  }
\label{fig:skill_based_generation_examples} 
\end{figure}

\begin{figure}[h]
  \centering
  \includegraphics[width=.95\linewidth]{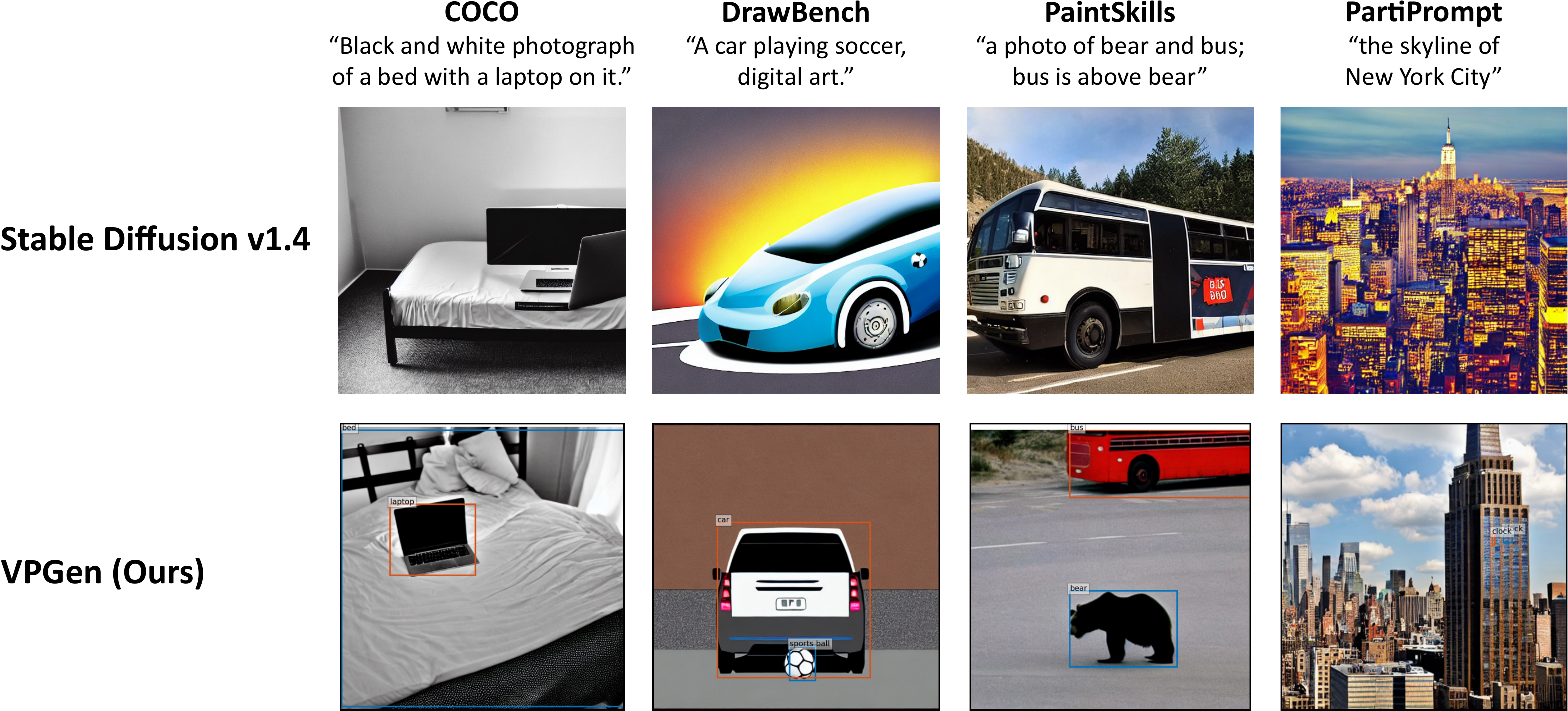}
  \caption{
    Images generated by \sdcompvis{} and our \textboxmethod{} (\vicuna{} 13B+\gligen{}) for open-ended prompts.
  }
\label{fig:open_ended_generation_examples} 
\end{figure}

\begin{figure}[h]
  \centering
  \includegraphics[width=.99\linewidth]{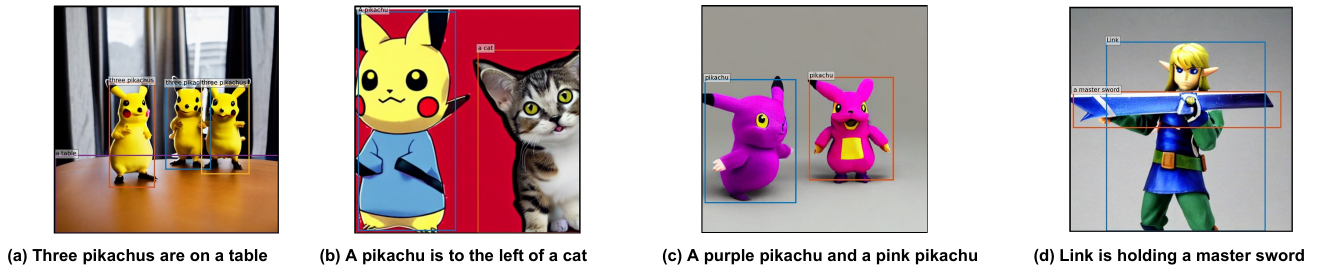}
  \caption{
    Images generated by \textboxmethod{} (\vicuna{} 13B+\gligen{}) from prompts including objects (\eg{} `Pikachu', `Link', `master sword') that are not included in text-to-layout annotations (Flickr30K Entities, COCO, PaintSkills) for \vicuna{} fine-tuning. 
  }
\label{fig:unseen_object_examples} 
\end{figure}

In \cref{fig:skill_based_generation_examples} and \cref{fig:open_ended_generation_examples},
we provide example images generated by our \textboxmethod{} and \sdcompvis{} (the closest baseline of \textboxmethod{}) with 
various skill-based and open-ended prompts, respectively.
In \cref{fig:unseen_object_examples}, we show \textboxmethod{} generation examples of placing objects that are unseen during the text-to-layout finetuning of \vicuna{} 13B. 

\section{Additional \textboxmethod{} Analysis}
\label{sec:gen_model_additional_experiments}

In the following, we provide additional analysis of \textboxmethod{}, including counting prompts, bounding box quantization, using GPT-3.5-Turbo as layout generation LM, and visual quality metric (FID).

\begin{figure}[h]
  \centering
  \includegraphics[width=.99\linewidth]{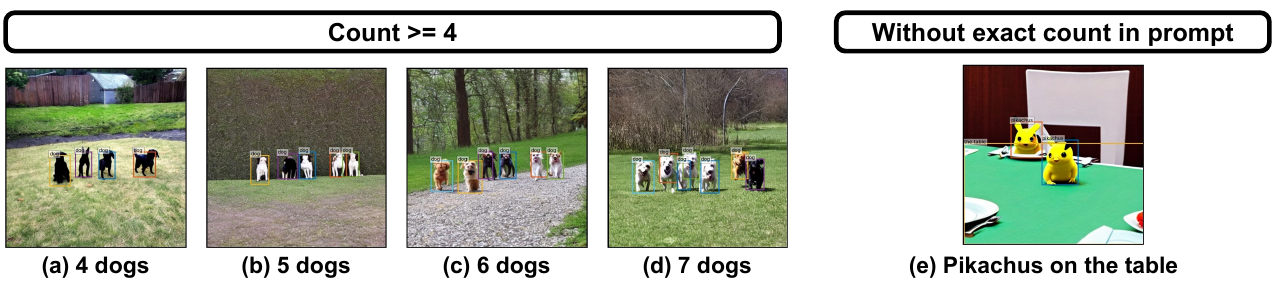}
  \caption{Images generated by our \textboxmethod{} with prompts requiring counting skills.
  (a) - (d) show the generation examples with count $\ge$ 4.
  (e) shows generation from a prompt that does include exact count (`pikachus').
  }
\label{fig:counting_examples} 
\end{figure}

\paragraph{Counting prompts: More than 4 objects \& Unspecified number of objects.}

In addition to the \countskill{} skill results with counts of 1 to 4 in the main paper \cref{fig:fine_grained_skills},
we provide additional results of \textboxmethod{} with counts of up to 7 in \Cref{tab:vpgen_count_performance} and show generation examples in \cref{fig:counting_examples} (a) - (d).
As expected, the trend continues from \cref{fig:fine_grained_skills}, where higher counts are more challenging. 
We also experiment with prompts without specific numbers. We find that \textboxmethod{} can successfully generate layouts/images even when the prompts do not have explicit counts. As shown in \cref{fig:counting_examples} (e), our \textboxmethod{} generates two Pikachus from the prompt ``Pikachus on the table''.

\begin{table}[h]
  \caption{\eval{} \countskill{} skill accuracy of \textboxmethod{} with different number of objects.
  }
  \label{tab:vpgen_count_performance}
  \centering
  \resizebox{0.5\columnwidth}{!}{
      \begin{tabular}{l cccccccc}
        \toprule
         & \multicolumn{7}{c}{VPEval \countskill{} Acc. (\%) $\uparrow$} \\
         \cmidrule(r){2-8}
         & 1 & 2 & 3 & 4 & 5 & 6 & 7 \\
        \midrule
        \textboxmethod{} & 85.3 & 80.1 & 65.6 & 55.2 & 48.0 & 35.3 & 28.1 \\
        \bottomrule
      \end{tabular}
  }
\end{table}

\paragraph{Bounding box quantization.}

As mentioned in the main paper, we use normalized and quantized coordinates with 100 bins to represent bounding boxes, following LM-based object detection works~\cite{Chen2022Pix2Seq,Yang2022UniTAB}.
In \Cref{tab:vpgen_bbox_ablation}, we compare 100 and 1000 bins for the bounding boxes. Increasing the granularity of the bounding box to 1000 does not increase the accuracy in both skill-based and open-ended prompts, suggesting that the current 100-bin quantization does not hurt the accuracy of object placements.

\begin{table}[h]
\centering
  \caption{Bounding box quantization: 1000 \vs{} 100 bins.
  }
  \label{tab:vpgen_bbox_ablation}
  \resizebox{0.7\columnwidth}{!} {
      \begin{tabular}{l c c }
        \toprule
        \multirow{1}{*}{\# Bins} & VPEval skill-based Acc. (\%) $\uparrow$ & VPEval open-ended Acc.  (\%) $\uparrow$\\
        \midrule
        1000 & 49.8 & 68.2 \\
        100 (default) & \textbf{51.0} & \textbf{68.3} \\
        \bottomrule
      \end{tabular}
  }
\end{table}

\paragraph{GPT-3.5-Turbo \vs{} Vicuna 13B.}

In our initial experiment, we tested GPT-3.5-Turbo to generate spatial layouts by showing in-context examples and found that the generated layouts were often inaccurate or not meaningful. For example, GPT-3.5-Turbo often generates a list of bounding boxes that is the same size as the entire image (\eg{} \texttt{[object 1 (0,0,99,99), object 2 (0,0,99,99) object3 (0,0,99,99)]}). We conjecture that this is because their training corpus might not include many bounding boxes. Then, we collected text-layout annotations and trained a language model to generate spatial layouts from text prompts.

For a quantitative comparison, we implement \textboxmethod{} with GPT-3.5-turbo with 36 in-context examples that cover different skills and compare it with \vicuna{} 13B trained on Flickr30k+COCO+PaintSkills. As shown in \Cref{tab:vpgen_lm_ablation}, \vicuna{} 13B based \textboxmethod{} shows higher skill-based and open-ended VPEval accuracies than GPT-3.5-Turbo based \textboxmethod{}.

\begin{table}[h]
\centering
  \caption{GPT-3.5-Turbo \vs{} \vicuna{} 13B in \textboxmethod{}.
  }
  \label{tab:vpgen_lm_ablation}
  \resizebox{0.95\columnwidth}{!} {
      \begin{tabular}{l c c }
        \toprule
        \multirow{1}{*}{Model} & VPEval skill-based Acc.  (\%) $\uparrow$ & VPEval open-ended Acc.  (\%) $\uparrow$\\
        \midrule
        GPT-3.5-Turbo (36 examples) + \gligen{} & 40.7 & 65.5 \\
        \vicuna{} 13B + \gligen{} & \textbf{51.0} & \textbf{68.3} \\
        \bottomrule
      \end{tabular}
  }
\end{table}

\paragraph{Visual quality metric (FID).}
In \Cref{tab:fid_and_accuracy},
we compare our \textboxmethod{} (\vicuna 13B + \gligen{}) to its backbone \sdcompvis{},
in FID~\cite{Heusel2017} (30K images of COCO val 2014 split) and VPEval Accuracy.
Both \textboxmethod{} checkpoints show better skill-based accuracy than \sdcompvis{} while achieving comparable open-ended accuracy. and FID.
In FID (lower the better), we find \textboxmethod{} (Flickr30k) < \sdcompvis{} <  \textboxmethod{} (Flickr30k+COCO+Paintskills).
We think that a bit of increase (but still reasonably good) in the FID of the Flickr30k+COCO+Paintskills checkpoint is because the layouts of PaintSkills are different from those of natural scenes (COCO and Flickr30k).

\begin{table}[h]
  \caption{VPEval accuracy and FID of VPGen (\vicuna{} 13B + \gligen{}) and its closest baseline \sdcompvis{}.
  \textit{F30: Flickr30K Entities, C: COCO, P: PaintSkills}.
  }
  \label{tab:fid_and_accuracy}
  \centering
  \resizebox{.95\columnwidth}{!}{
      \begin{tabular}{l ccc}
        \toprule
        Model & VPEval Skill-based Acc. (\%) $\uparrow$ & VPEval Open-ended Acc. (\%) $\uparrow$ & FID (COCO 30K) $\downarrow$ \\
        \midrule
        \sdcompvis{} & 37.7 & 70.6 & 16.5 \\
        \midrule
        \textboxmethod{} (F30) & 43.9 & \textbf{71.0} & \textbf{15.9} \\
        \textboxmethod{} (F30+C+P) & \textbf{51.0} & 68.3 & 20.1 \\
        \bottomrule
      \end{tabular}
  }
\end{table}

\paragraph{Training with real \vs{} synthetic images.}
Our layout generation module (\vicuna{} 13B) in \textboxmethod{} is trained on a mix of both 3D simulator-based synthetic (PaintSkills) and real-world images (MSCOCO/Flickr30K).
We conduct an ablation study of training only with real \vs{} synthetic images.
We train the \vicuna{} on only PaintSkills and only Flickr30K and compare their results on both skill-based and open-ended prompts in \Cref{tab:vpgen_data_ablation}.
For skill-based prompts, PaintSkills training shows slightly higher average accuracy (46.0) than Flickr30K training (43.9).
For open-ended prompts, PaintSkills training gets a VPEval score of (65.6), lower than Flickr30K training (71.0).
This indicates that training on synthetic data can inject knowledge of specific skills such as counting and spatial relation understanding, but it is hard to cover diverse objects and attributes only with synthetic data.

\begin{table}[h]
\centering
  \caption{Training \vicuna{} 13B with real \vs{} synthetic data.
  \textit{F30: Flickr30K Entities, P: PaintSkills}.
  }
  \label{tab:vpgen_data_ablation}
  \resizebox{0.8\columnwidth}{!} {
      \begin{tabular}{l c c }
        \toprule
        \multirow{1}{*}{Model} & VPEval skill-based Acc. (\%) $\uparrow$ & VPEval open-ended Acc. (\%) $\uparrow$\\
        \midrule
        \textboxmethod{} (F30; real) & 43.9 & \textbf{71.0} \\
        \textboxmethod{} (P; synthetic) & \textbf{46.0} & 65.6 \\
        \bottomrule
      \end{tabular}
  }
\end{table}

\section{\textboxmethod{} Error Analysis}
\label{sec:vpgen_error_analysis}

\begin{table}
  \caption{Step-wise error analysis of VPGen (\vicuna{}+\gligen{}).
  For the \spatialskill{} skill, `front' and `behind' splits are skipped in this table since our Vicuna 13B does not generate depth information.
  }
  \label{tab:vpgen_error_prop}
  \centering
  \resizebox{0.95\columnwidth}{!}{
      \begin{tabular}{l cccc}
        \toprule
        \multirow[c]{2}{*}{Skills}& \multicolumn{3}{c}{\vicuna{} 13B} & \gligen{} \\
        \cmidrule{2-4} \cmidrule {5-5}
        & Object Recall (\%) $\uparrow$ & Object Count (\%) $\uparrow$ & Layout Accuracy (\%) $\uparrow$ & Image Accuracy (\%) $\uparrow$ \\
        \midrule
        Object & 99.2 & 98.8 & 99.2 & 96.8 \\
        Count & 99.1 & 98.8 & 98.8 & 72.2 \\
        Spatial & 98.3 & 98.2 & 87.5 & 63.3 \\
        Scale & 92.8 & 92.6 & 38.2 & 26.3 \\
        \bottomrule
      \end{tabular}
  }
  \vspace{-10pt}
\end{table}

\paragraph{Layout \vs{} Image generation on skill-based prompts.}
We analyze the performance of each step in the \textboxmethod{} pipeline to determine how much error is propagated with the following four metrics.
\begin{enumerate}
    \item Object Recall: The ratio of whether correct objects are included in the generated layouts.
    \item Object Count: The ratio of the correct number of objects is included in the generated layouts.
    \item Layout Accuracy: VPEval accuracy of the generated layouts (using the generated layouts as the object detection results).
    \item Image Accuracy: VPEval accuracy of the final image by \gligen{} (in the main Table 1).
\end{enumerate}
As shown in \Cref{tab:vpgen_error_prop}, the layout accuracy is much higher than image accuracy, especially for \countskill{}/\spatialskill{} skills.
This indicates that the major error source is the image rendering step (with \gligen{}), and using a more accurate layout-to-image generation model in the future would improve the accuracy of our \textboxmethod{} pipeline.

\paragraph{Human evaluation on open-ended prompts.}
As mentioned in the main paper \cref{sec:eval_results_open}, we ask two expert human annotators to evaluate 50 randomly sampled prompts from \tifaprompts{}~\cite{hu2023tifa}, with the following two metrics:
(1) layout accuracy: if the generated layouts align with the text prompts;
(2) image accuracy: if the generated images align with the text prompts.
The human annotators show high inter-annotator agreements for both layout accuracy (Cohen's $\kappa=0.73$,  Krippendorff's $\alpha=0.73$) and image accuracy (Cohen's $\kappa=0.87$,  Krippendorff's $\alpha=0.87$).
The \vicuna{} 13B's layout accuracy is (92\%) higher than \gligen{}'s image correctness (65\%).
The result is consistent with the skill-based prompts error analysis result (\Cref{tab:vpgen_error_prop}) and suggests that a better layout-to-image generation model could further improve our \textboxmethod{} framework.

\begin{figure}[h]
  \centering
  \includegraphics[width=.99\linewidth]{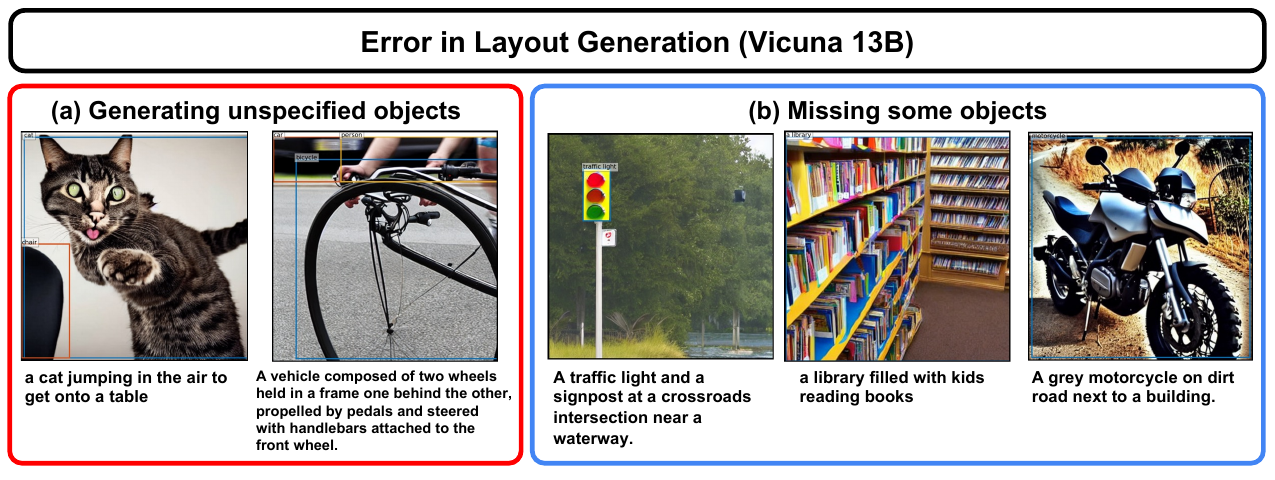}
  \caption{
  Text-to-Layout generation error examples.
  The \vicuna{} 13B sometimes (a) generates objects that are not specified in prompts (\eg{} `chair' in the 1st example) or (b) misses objects that are mentioned in prompts (\eg{} `building' in the 5th example).
  }
\label{fig:vpgen_layout_error} 
\end{figure}

\begin{figure}[h]
  \centering
  \includegraphics[width=.99\linewidth]{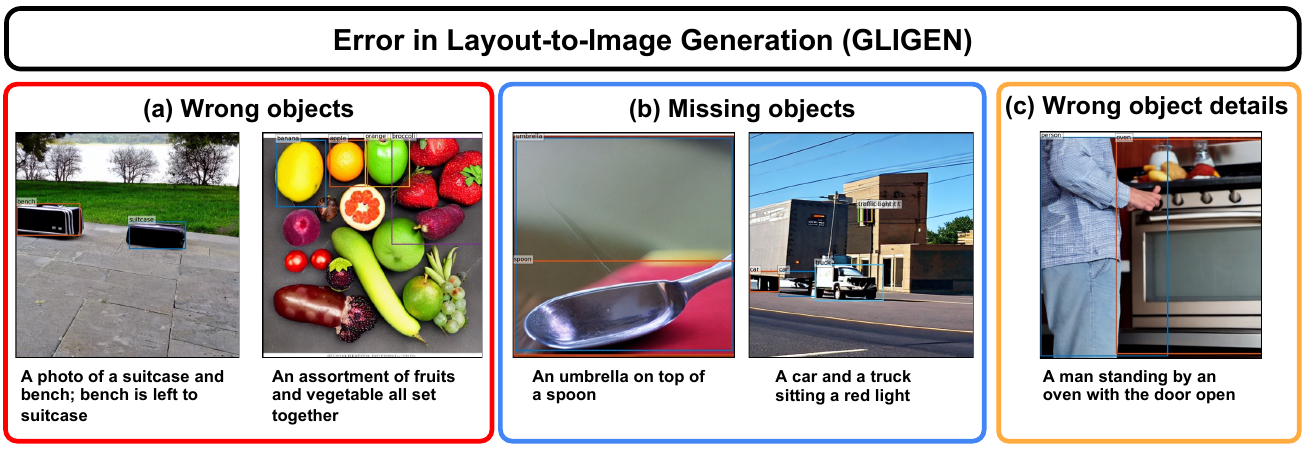}
  \caption{Layout-to-Image generation error examples.
  The \gligen{} sometimes (a) generates wrong objects (\eg{} `suitcase', instead of `bench', in the 1st example), (b) misses some objects (\eg{} `umbrella' in the 3rd example), or (c) misses object details (\eg{} `closed oven' instead of `oven with the door open' in the 5th example).
  }
\label{fig:vpgen_gligen_error} 
\end{figure}

\paragraph{Common error categories.}
We show some common categories of errors from the text-to-layout generation step (with \vicuna{} 13B) and from the layout-to-image generation step (with \gligen{}), respectively.
As shown in \cref{fig:vpgen_layout_error},
the \vicuna{} 13B sometimes (a) generates objects that are not specified in prompts (\eg{} `chair' in the 1st example) or (b) misses objects that are mentioned in prompts (\eg{} `building' in the 5th example).
As shown in \cref{fig:vpgen_gligen_error},
the \gligen{} sometimes (a) generates wrong objects (\eg{} `suitcase', instead of `bench', in the 1st example), (b) misses some objects (\eg{} `umbrella' in the 3rd example), or (c) misses object details (\eg{} `closed oven' instead of `oven with the door open' in the 5th example).

\begin{figure}[h]
  \centering
  \includegraphics[width=.8\linewidth]{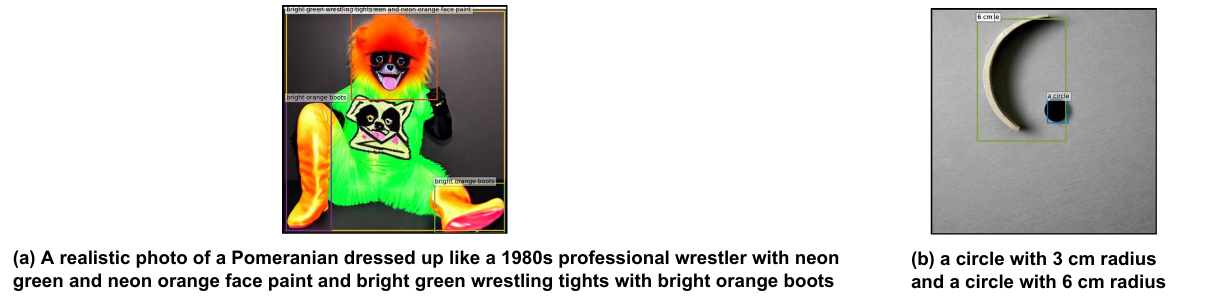}
  \caption{
    Images generated by \textboxmethod{} (\vicuna{}+\gligen{}) from challenging, non-canonical prompts.
  }
\label{fig:challenging_prompts_examples} 
\end{figure}

\paragraph{Challenging, non-canonical prompts.}
In \cref{fig:challenging_prompts_examples}, we show generation examples with challenging, non-canonical prompts (a) ``A realistic photo of a Pomeranian dressed up like a 1980s professional wrestler with neon green and neon orange face paint and bright green wrestling tights with bright orange boots'' (from DrawBench~\cite{Saharia2022Imagen}) and (b) ``a circle with 3 cm radius and a circle with 6 cm radius''.
Our \textboxmethod{} understands the important parts of the prompts (\eg{} generating `realistic', `Pomeranian dog', `bright orange boots' and generating layouts of two circles in different sizes), but misses some aspects (\eg{} the bigger circle is not twice the size of the smaller circle).
This is probably because the training prompts (COCO, PaintSkills, Flickr30k Entities) do not include many prompts written in these styles. We believe that scaling datasets with various sources can further improve \textboxmethod{}.

\section{\eval{} Error Analysis}
\label{sec:vpeval_error_analysis}

\paragraph{Evaluation program analysis: coverage and accuracy.}

\begin{figure}
  \centering
  \includegraphics[width=.95\linewidth]{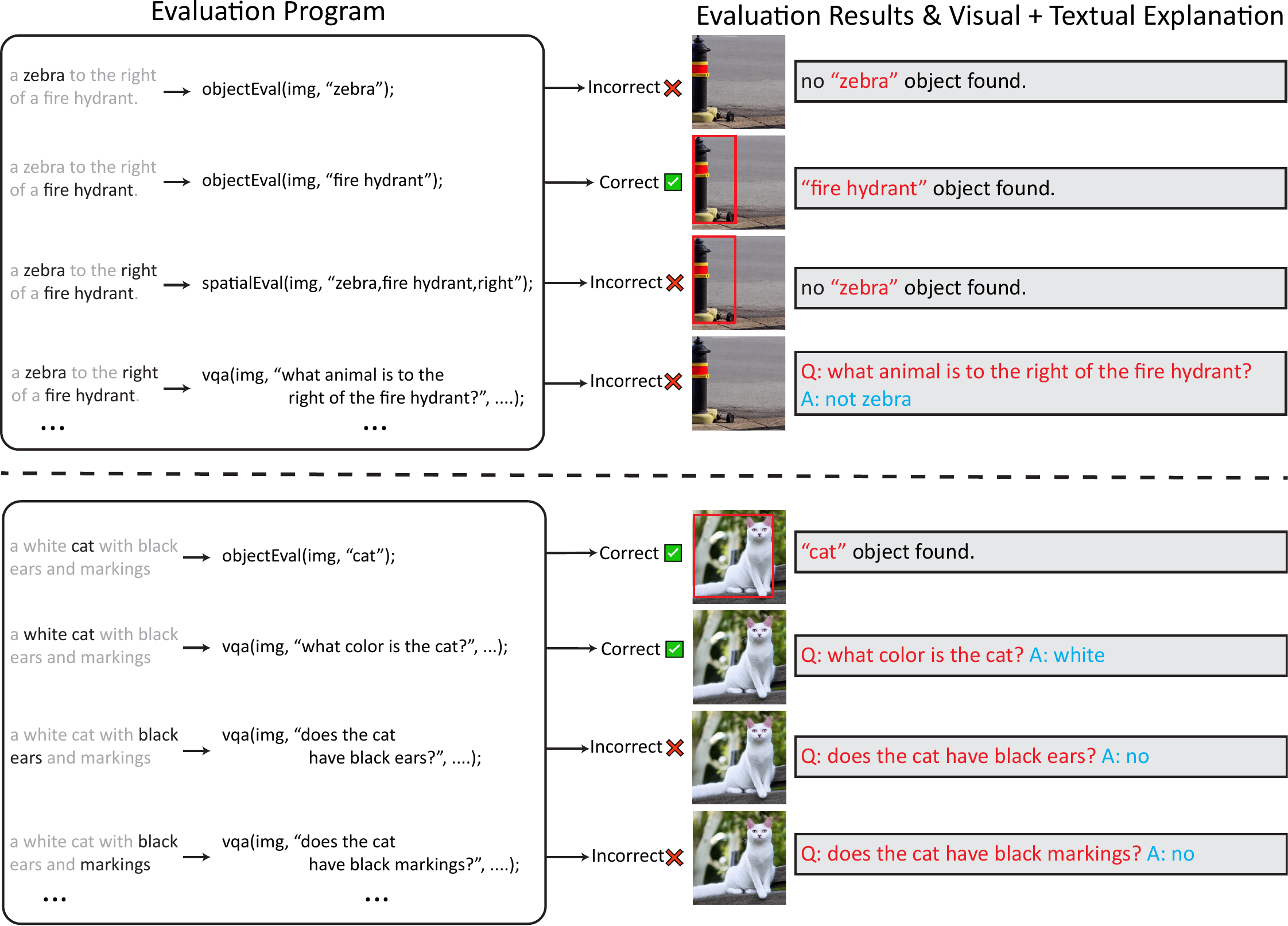}
  \caption{
  Examples of our \eval{} evaluation programs and module outputs.
  Each module covers different parts of the text prompts (see \cref{sec:vpeval_error_analysis} for discussion).
  }
\label{fig:vpeval_analysis} 
\end{figure}

For the evaluation program analysis (main paper \cref{sec:human_correlation}),
we experiment with two expert annotators using 50 randomly sampled \tifaprompts{}.
We ask two expert annotators to analyze the evaluation programs in two criteria:
(1) how well the evaluation program covers the content from the text prompt;
(2) how accurately each evaluation module evaluates the images.
To make it easier for the annotators, for (2) module accuracy analysis, we only ask them to check the output of three randomly sampled evaluation modules.
We find that, on average, our evaluation programs cover 94\% of the prompt elements, and our modules are accurate 83\% of the time.
This shows that our \eval{} evaluation programs are effective in selecting what to evaluate and are highly accurate while also providing interpretability and explainability.
In \cref{fig:vpeval_analysis},
we visualize our evaluation programs that cover different parts of the prompts and evaluation modules that provide accurate outputs and visual+textual explanations.

\section{\textboxmethod{}/\eval{} Implementation Details}
\label{sec:implemenation_details}

\subsection{\textboxmethod{} Details}

\noindent\textbf{Training layout-aware LM.}
To obtain layout-aware LM,
we use Vicuna 13B~\cite{vicuna2023},
a public
state-of-the-art
chatbot language model finetuned from LLaMA~\cite{Touvron2023LLaMA}.
We use parameter-efficient finetuning with LoRA~\cite{Hu2022LoRA} to preserve the original knowledge of LM and save memory during training and inference.
We collect the text-layout pair annotations from training sets of three public datasets:
Flickr30K entities~\cite{Plummer2017Flickr30kEntities}, MS COCO instances 2014~\cite{Lin2014COCO}, and PaintSkills~\cite{Cho2023DallEval}, totaling 1.2M examples.
We use randomly selected 2,000 examples for the validation set and use the rest for training.
For the object/count generation step,
we set the maximum counts for a single object class as 7.
We train \vicuna{} 13B with per-gpu batch size 96 (= 24 batch x 4 gradient accumulation).
When training \vicuna{} 13B with Flickr30K+COCO+PaintSkills dataset,
we train the model for 2 epochs.
When training only with Flickr30K dataset,
we train the model for 6 epochs, to rough match the training time.
Training takes 26 hours with 4 A6000 GPUs (each 48GB).

\noindent\textbf{\gligen{} inference details.}
We use Huggingface Diffusers~\cite{von-platen-etal-2022-diffusers} implementation of \gligen{}~\cite{Li2023GLIGEN}\footnote{\url{https://github.com/gligen/diffusers}}.
Following the default configuration, we use $\texttt{gligen\_scheduled\_sampling\_beta}=0.3$, $\texttt{num\_inference\_steps}=50$, and fp16 precision during inference.

\paragraph{Inference time.}
On a single A6000 GPU, \textboxmethod{} takes around 6s (2s for Vicuna 13B and 4s for GLIGEN) to generate an image.

\subsection{\eval{} Details}
\label{sec:vpeval_details}

\paragraph{\eval{} prompts and evaluation programs.}

Our skill-based benchmark has an almost uniform distribution of objects and relations among the text prompts.
This ensures that generation models cannot achieve high scores by performing highly on a few common objects, relations, and counts, \etc{}.
We create skill-specific prompts by composing templates with lists of objects (from 80 COCO objects), counts (1-4), relations (2D: above, below, left, right; 3D: front, behind), scale (smaller, bigger, same), and text (31 unique words).
Our evaluation can be extended to any number of objects/counts/relations/\etc{}.
In total, our skill-based benchmark has 400/1000/1000/1000/403 prompts for the object/count/spatial/size/text skills, respectively.
In \Cref{tab:full_prompt_list}, we show templates and evaluation programs used for each skill.

\begin{table}[t]
    \caption{The prompt templates and evaluation programs for skill-based evaluation. We also insert grammar tokens like \texttt{<a>} that are replaced with ``an'' or ``a'' to maintain correct grammar.}
    \label{tab:full_prompt_list}
    \centering
    \resizebox{0.95\linewidth}{!} {
        \begin{tabular}{l | c | c | c | c | c}
            \toprule
            Skill & \objectskill{} & \countskill{} & \spatialskill{} & \scaleskill{} & \textskill{} \\
            \midrule
            Prompt & \makecell {
                \texttt{<objA>} \\
                \texttt{<a>} \texttt{<objA>} \\
                a photo of \texttt{<a>} \texttt{<objA>} \\
                an image of \texttt{<a>} \texttt{<objA>} \\
                a picture of \texttt{<a>} \texttt{<objA>}
            } &
                    \makecell{
            \texttt{<N>} \texttt{<objA>}\texttt{<s>} \\
            a photo of \texttt{<N>} \texttt{<objA>}\texttt{<s>} \\
            a picture of \texttt{<N>} \texttt{<objA>}\texttt{<s>} \\
            an image of \texttt{<N>} \texttt{<objA>}\texttt{<s>} \\
            \texttt{<N EN>} \texttt{<objA>}\texttt{<s>} \\
            a photo of \texttt{<N EN>} \texttt{<objA>}\texttt{<s>} \\
            a picture of \texttt{<N EN>} \texttt{<objA>}\texttt{<s>} \\
            an image of \texttt{<N EN>} \texttt{<objA>}\texttt{<s>} \\
        } &
        \makecell{
         \texttt{<a2>} \texttt{<objB>} is \texttt{<tothe>} \\ \texttt{<rel>}\texttt{<of>} \texttt{<a1>} \texttt{<objA>} \\
        } &
        \makecell{
         \texttt{<a2>} \texttt{<objB>} that is \\ \texttt{<scale>} than \texttt{<a1>} \texttt{<objA>} \\
        } &
        \makecell{
         a sign that reads '\texttt{<text>}' \\
a book cover that reads '\texttt{<text>}' \\
a poster that reads '\texttt{<text>}' \\
a sign that says '\texttt{<text>}' \\
a book cover that says '\texttt{<text>}' \\
a poster that says '\texttt{<text>}' \\
a storefront with '\texttt{<text>}' written on it \\
a storefront with '\texttt{<text>}' written \\
a storefront with '\texttt{<text>}' displayed \\
a piece of paper that says '\texttt{<text>}' \\
a piece of paper that reads '\texttt{<text>}' \\
a piece of paper that says '\texttt{<text>}' on it \\
a piece of paper that reads '\texttt{<text>}' on it \\
        } \\
        \midrule

        Evaluation Program &
        objectEval(img, <objA>) &
        countEval(img, <objA>, <N>) &
        spatialEval(img, <objA>,<objB>,<rel>)  &
        scaleEval(img, <objA>,<objB>,<scale>) & 
        textEval(img, <text>) \\

        \bottomrule
        \end{tabular}
    }
\end{table}

\paragraph{Captioning/VQA/CLIP baseline details.}
For captioning, we generate a caption with BLIP-2~\cite{Li2023BLIP2} and calculate text similarity metrics between the caption and the original text prompt.
For VQA, we give the BLIP-2 (Flan-T5 XL) the image, a question, and a yes/no answer choice, with a prompt template \texttt{``Question: \{question\} Choices: yes, no Answer:''}. The \texttt{\{question\}} is a formatted version of the text prompt (\eg{} ``a photo of a dog'' $\rightarrow$ ``is there a dog in the image?''). For the spatial and scale skill, we ask three questions:
if object A is present (\eg{} ``is there a dog in the image?'')
if object B is present (\eg{} ``is there a cat in the image?''),
and if the relationship between them is true (\eg{} ``is the cat behind the dog?'').
We mark each image as correct if the VQA model outputs `yes' for all three questions.
The first two questions prevent
accidental false positives if an object is missing.
For CLIP (ViT-B/32)~\cite{Radford2021CLIP},
we encode the generated image and text prompts,
and 
take the cosine similarity between the visual and text features as the final score. 

\paragraph{Inference time.}
On a single A6000 GPU, each \eval{} module typically takes less than 1s to run.

\paragraph{Program generation API cost.}

Using GPT-3.5 Turbo\footnote{\url{https://openai.com/pricing}} as the program generator,
\eval{} takes about \$0.007 (less than 1 cent) to generate programs from a prompt (4K input tokens, 0.5K output tokens).

\section{TIFA Prompt Element Analysis Details}
\label{sec:tifa_prompt_element_analysis_details}

To better understand the characteristics of the TIFA prompts~\cite{hu2023tifa} that we use in open-ended evaluation,
we randomly sample 20 prompts and label whether each prompt has any elements that require an understanding of spatial layouts, as mentioned in the main paper
\cref{sec:eval_results_open}.
Each prompt has multiple elements about objects, attributes, spatial layouts, \etc{}.
For example, ``a zebra to the right of a fire hydrant.'' has the elements: `zebra', `right', and `fire hydrant'.
From the 20 prompts, there are 81 total elements, and only 11/81 (13.6\%) are relevant to object layouts (\eg{} `right'). The remaining 70 elements are all related to objects or attributes (\eg{} ``a white cat with black ears and markings'' has the elements `cat', `white cat', `ears', `black ears', `markings', and `black markings').

\section{Limitation and Broader Impacts}
\label{sec:limitation}

\paragraph{Limitations.}

Since our LMs were primarily trained with English-heavy datasets, they might not well understand prompts written in non-English languages.
Likewise, as our generation/evaluation modules were primarily trained with natural images, they might not well handle images in different domains (\eg{} fine-arts, medical images, \etc{}).
Note that our framework supports easy extensions so that users can update existing modules or add new ones with better domain-specific knowledge.

Generating evaluation programs using LLMs can be costly both in terms of computational resources and price (API cost).
Thus, we release the evaluation programs so that users do not have to re-generate the evaluation programs. In addition, we will release a public LM (finetuned for evaluation program generation using ChatGPT outputs) that can run on local machines.

\paragraph{Broader Impacts.}

While our interpretable \textboxmethod{} framework can be beneficial to many applications (\eg{} user-controlled image generation/manipulation and data augmentation), it could also be used for potentially harmful cases (\eg{} creating false information or misleading images), like other image generation frameworks.

\section{License}
\label{sec:license}

We will make our code and models publicly accessible. 
We use standard licenses from the community and provide the following links to the licenses for the datasets, codes, and models that we used in this paper. 
For further information, please refer to the links.

\vspace{3pt}
\noindent\textbf{FastChat (Vicuna):} \href{https://github.com/lm-sys/FastChat/blob/main/LICENSE}{Apache License 2.0}

\vspace{3pt}
\noindent\textbf{GLIGEN:} \href{https://github.com/gligen/GLIGEN/blob/master/LICENSE}{MIT}

\vspace{3pt}
\noindent\textbf{Grounding DINO:} \href{https://github.com/IDEA-Research/GroundingDINO/blob/main/LICENSE}{Apache License 2.0}

\vspace{3pt}
\noindent\textbf{DPT:} \href{https://github.com/isl-org/DPT/blob/main/LICENSE}{MIT}

\vspace{3pt}
\noindent\textbf{LAVIS (BLIP-2):} \href{https://github.com/salesforce/LAVIS/blob/main/LICENSE.txt}{BSD 3-Clause}

\vspace{3pt}
\noindent\textbf{TIFA:} \href{https://github.com/Yushi-Hu/tifa/blob/main/LICENSE}{Apache License 2.0}

\vspace{3pt}
\noindent\textbf{\mindalle{}:} \href{https://github.com/kakaobrain/mindall-e/blob/main/LICENSE}{Apache License 2.0, CC-BY-NC-SA 4.0}

\vspace{3pt}
\noindent\textbf{DALL-E mini (\dallemega{}):} \href{https://github.com/borisdayma/dalle-mini/blob/main/LICENSE}{Apache License 2.0}, \href{https://github.com/kuprel/min-dalle/blob/main/LICENSE}{MIT}

\vspace{3pt}
\noindent\textbf{EasyOCR:} \href{https://github.com/JaidedAI/EasyOCR/blob/master/LICENSE}{Apache License 2.0}

\vspace{3pt}
\noindent\textbf{OpenAI API (ChatGPT):} \href{https://github.com/openai/openai-openapi/blob/master/LICENSE}{MIT}

\vspace{3pt}
\noindent\textbf{Diffusers:} \href{https://github.com/huggingface/diffusers/blob/main/LICENSE}{Apache License 2.0}

\vspace{3pt}
\noindent\textbf{Transformers:} \href{https://github.com/huggingface/transformers/blob/main/LICENSE}{Apache License 2.0}

\vspace{3pt}
\noindent\textbf{PyTorch:} \href{https://github.com/pytorch/pytorch/blob/main/LICENSE}{BSD-style}

\end{document}